\documentclass{article}

\PassOptionsToPackage{numbers}{natbib}

\usepackage[main, final]{neurips_2026}

\usepackage[utf8]{inputenc} 
\usepackage[T1]{fontenc}

\usepackage{amsmath}
\usepackage{amsfonts}
\usepackage{amssymb}
\usepackage{nicefrac}

\usepackage[dvipsnames, table]{xcolor} % Load once with all options
\usepackage{graphicx}
\usepackage{rotating} % For sidewaystable

\usepackage{array}
\usepackage{booktabs}
\usepackage{multirow}
\usepackage{caption}
\usepackage{subcaption}
\usepackage{wrapfig}
\usepackage{microtype}
\usepackage{url}
\usepackage{subcaption}
\usepackage{algorithm}
\usepackage{algpseudocode}
\usepackage{xcolor}

\usepackage{hyperref} 

\usepackage{cleveref}

\newcolumntype{x}[1]{>{\centering\arraybackslash}p{#1}}
\newcommand{\tablestyle}[2]{\setlength{\tabcolsep}{#1}\renewcommand{\arraystretch}{#2}\centering\fontsize{8pt}{9pt}\selectfont}
\definecolor{dark_red}{HTML}{d62728} % Using Indian Red
\newcommand{\authorskip}{\hspace{2.5mm}}
\newcommand{\institutionskip}{\hspace{5.0mm}}

\title{Invaria: Learning Scale and Density Invariance in Point Clouds via Next-Resolution Prediction}

\author{Chun-Peng Chang\textsuperscript{\mdseries1} \authorskip 
Shaoxiang Wang\textsuperscript{\mdseries2} \authorskip
Alain Pagani\textsuperscript{\mdseries3} \authorskip
Dariu Gavrila\textsuperscript{\mdseries1} \authorskip
Holger Caesar\textsuperscript{\mdseries1} \authorskip \\ \\
\textsuperscript{1}TU Delft, Cognitive Robotics \institutionskip
\textsuperscript{2}RPTU \institutionskip
\textsuperscript{3}DFKI, Augmented Vision
}

\begin{document}
\maketitle
\begin{abstract}
Modern image encoders achieve high generalization by decoupling semantic meaning from resolution, an ability yet to be fully realized in the 3D domain. We investigate the failure of 3D point cloud encoders to achieve similar generalization and find that existing models are highly sensitive to sampling resolution and scale changes, leading to significant performance degradation. This sensitivity is a major bottleneck for real-world deployment in robotics, as it suggests models overfit to specific quantization densities and object scales rather than learning invariant semantic features. To mitigate this dependency, we propose Invaria, a point cloud encoder that achieves scale and density invariance through next-resolution prediction and receptive field calibration. While our objective is not the explicit generation of high-resolution point clouds, we find that this training objective encourages the model to learn robust, structural invariants. The resulting encoder achieves significant performance gains during resolution shifts while maintaining high efficiency through a compact model size and reduced token requirements. Specifically, on ScanNet, Invaria achieves a 56.0\% higher mIoU at 3$\times$ lower resolution and a 20\% improvement when the objects scale is reduced by a factor of 3. These gains are achieved with a 45\% smaller model size and an average reduction of 40\% in input tokens.
\url{https://birdy666.github.io/projects/invaria/}
\end{abstract}

\section{Introduction}
\label{sec:introduction}

Humans possess an inherent ability to recognize objects regardless of scale and resolution. Modern image encoders~\cite{radford2021learning, dosovitskiy2020image} have successfully mirrored this capability, enabling seamless domain transfer from high-quality public datasets to real-world or simulated environments without requiring retraining or fine-tuning.
In 3D point cloud, this ability is equally critical for deployment such as robotics when the device generates non-uniform and low resolution point clouds that vary with sensor quality and hardware/latency constraints~\cite{wang2020train}. 

\begin{figure}
  \centering
   \includegraphics[width=1\linewidth]{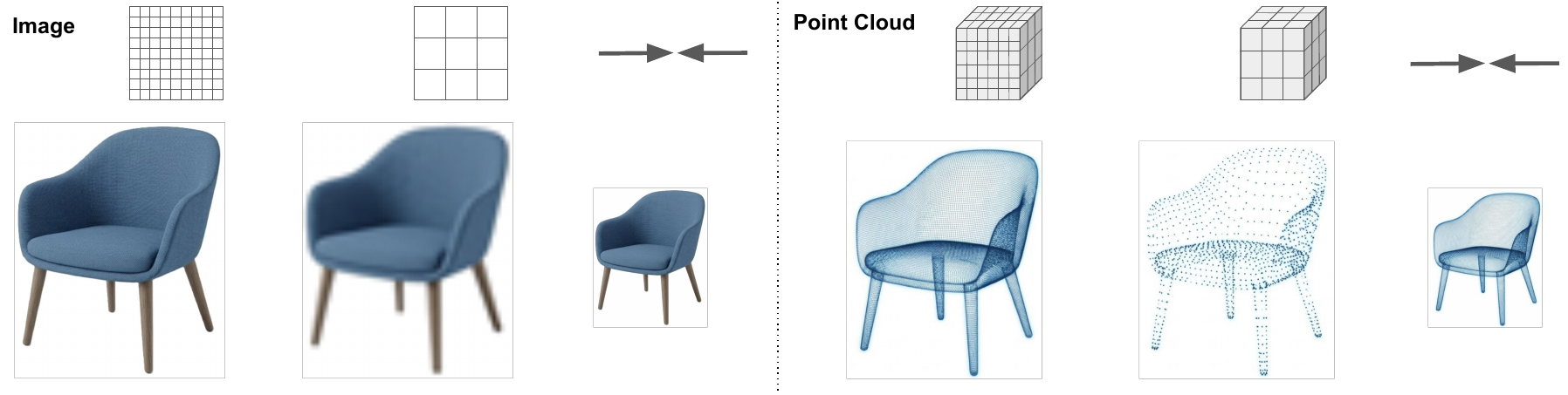}
   \caption{The generalization gap in 3D understanding. (Left) Humans and modern image encoders can seamlessly generalize a semantic concept (such as "chair") across diverse resolutions and sizes (Right). 
   Modern 3D encoders, on the other hand do not demonstrate similar ability, as they fail to understand semantic concepts independently of the underlying point density and scale.}
   \label{fig:chair}
\end{figure}

However, despite significant progress in 3D perception~\cite{choy20194d, wang2023octformer, zhao2021point, wu2022point, wu2025sonata, zhang2025concerto}, our research suggests that existing point cloud methods remain inherently dependent on specific sampling resolutions, which is often defined as the discretization grid size used to quantize local signals. This dependency causes performance degradation whenever the input resolution shifts from the training distribution. In this work we focus on two primary factors contributing to this sensitivity:
(1) \textbf{Density}: the number of points used to represent an object, which can affect the model's ability to recognize the object~\cite{qi2017pointnet++, komarichev2019cnn}; and 
(2) \textbf{Scale}: Operations within the model such as pooling and sparse convolution rely on the grid size to define their effective receptive field~\cite{wang2023octformer}; 
changing the grid size therefore also changes the object size perceived by the model~\cite{qian2022pointnext}. This highlights another deficiency in existing models: their failure to acquire scale-invariant features, which prevents reliable object recognition when the scale deviates from the training distribution. See \Cref{fig:chair} for the example.

To address these challenges, we propose \textit{Invaria}, a method that prevents models from depending on specific input resolutions and enhances robustness against density and scale changes. Existing encoders often struggle because they easily adopt "shortcuts" during training~\cite{geirhos2020shortcut, wu2025sonata}. Specifically, they learn to recognize objects based on a specific point density or a fixed receptive field defined by the grid size (scale), rather than the object's actual shape. 
Invaria mitigates this by utilizing a next-resolution prediction training pipeline with receptive field calibration. By predicting the semantic features of higher-resolution targets from sparser inputs in the latent space, the model learns to decouple semantic meaning from specific sampling resolutions. This training objective encourages the encoder to capture the object's fundamental structural invariants, the geometric relationships that define it regardless of resolution or scale.

Beyond robust feature representation, model efficiency, specifically in terms of parameter count and computational overhead, is a critical metric for 3D perception. Following the success of transformers~\cite{vaswani2017attention, beyer2023flexivit, carion2020end, zhu2020deformable}, state-of-the-art models~\cite{zhao2021point, wu2022point, wu2024point, liu2021swin} have shifted toward transformer-based architectures, which are powerful yet often computationally expensive. Furthermore, as self-supervised learning (SSL) and scaling laws have proven their efficacy, recent works~\cite{wu2025sonata, zhang2025concerto, zhang2026utonia} have leveraged SSL and increased model and data scales to enhance performance. While effective, this progress comes at a significant cost: the reliance on massive data and model sizes leads to high computational requirements, complicating real-world deployment on resource-constrained devices.

Conversely, by training with our proposed pipeline, our method demonstrates robust performance across varying resolutions while maintaining competitive performance at the default resolution with a significantly smaller model size. Specifically, when compared to state-of-the-art models on ScanNet, our method achieves a 56.0\% higher mIoU at 3$\times$ lower resolution (i.e., 3$\times$ the default grid size); and 20\% higher mIoU when the objects scale is reduced by a factor of 3. 
Moreover, this resolution-invariant property provides a significant computational advantage: the model can encode high-resolution features from lower-resolution inputs. This asymmetric encoding yields an average reduction of around 40\% in input tokens, which directly lowers the cost of transformer-based architectures, whose computation and memory grow with the number of input tokens. Consequently, our approach allows the network to maintain dense semantic representations while significantly lowering the computational overhead associated with the backbones.

In summary, the main contributions of this work are as follows:

\begin{itemize}
    \item  We investigate the sensitivity of existing point cloud encoders to sampling resolution and specifically focus on how the density shortcut and scale dependency can degrade the model's performance.  
    \item We introduce \textbf{Invaria}, a point cloud encoder designed to learn scale- and density-invariant features. Specifically, we apply next-resolution prediction with receptive field calibration to encourage the model to align features in the latent space that remain robust as resolution and scale shift.

    \item In contrast to current methods that rely on increasing data and model scales, Invaria achieves invariant features with a more compact architecture. Moreover, by leveraging resolution-invariant representations we can further reduce input token counts by utilizing lower-resolution point clouds, a crucial advantage for transformer-based backbones.
    
\end{itemize}

\section{Semantic Consistency and Resolution Shift}
\label{sec:problem}
\begin{figure}[t] % 't' helps placement at the top of the page
\captionsetup{skip=3pt}
   \centering
   \includegraphics[width=1\linewidth]{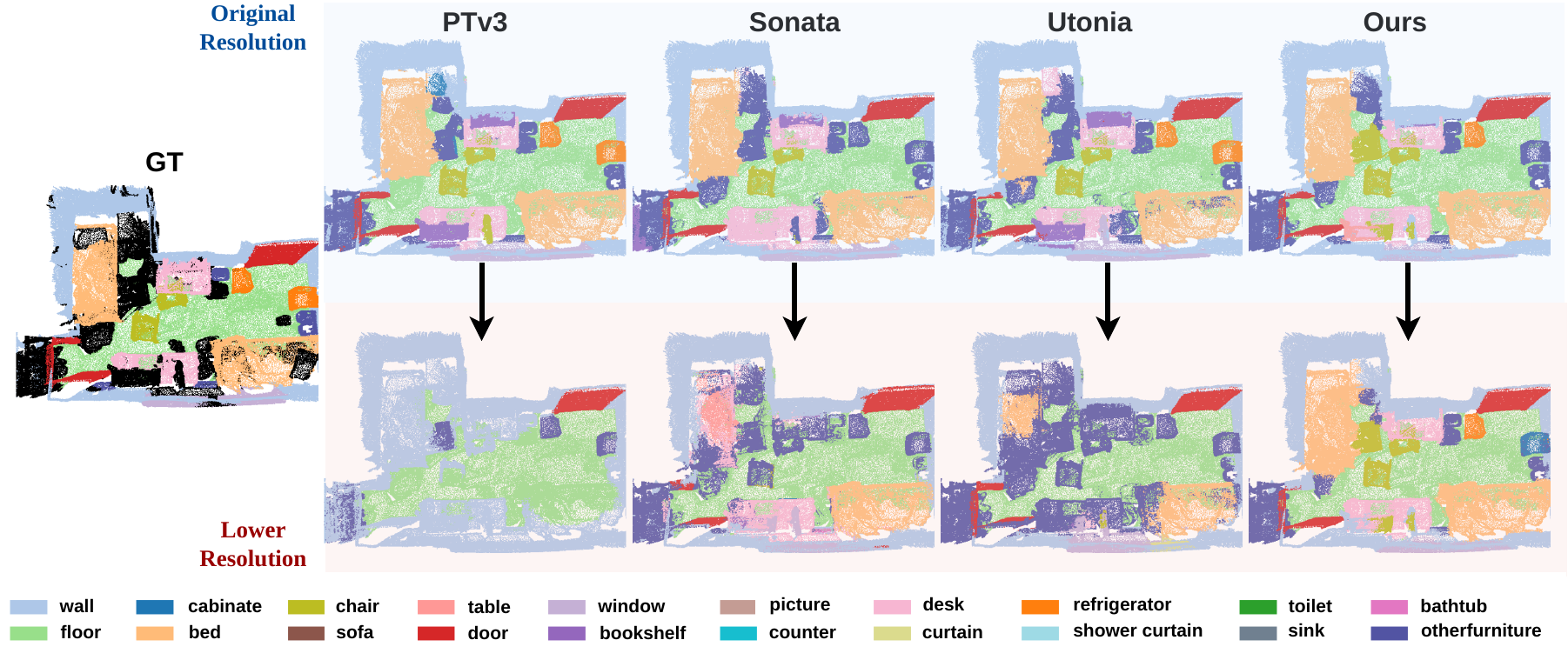}
   \caption{Comparison on ScanNet semantic segmentation. The first row displays model performance when evaluated at the same resolution. The second row illustrates the impact of shifting the resolution during inference. For clear visual comparison, results from both methods are projected on the original point cloud. The black spot in GT means no label. See \Cref{sec:more_vis} \Cref{fig:qualitative_more} for more examples.}
   \label{fig:qualitative}
   \vspace{-5mm}
\end{figure}

The representation of the continuous world in digital form necessitates a discretization process that reduces continuous visual detail into a discrete set of data points. 
In the 2D domain, this is achieved by sampling a scene at a specific resolution with corresponding number of pixels. 
Modern image encoders~\cite{caron2021emerging,kirillov2023segment,liu2021swin,xie2021segformer,ravi2024sam} have developed strong capabilities in extracting high-level abstract features that maintain semantic consistency across these variations. 
Similarly, in 3D, the resolution $R$ is usually defined by the grid size $\tau$ used for spatial quantization, which serves as a discretization threshold that quantizes local signals of the continuous scene $\mathcal{C}$ into a single representative point. This sampling process $\mathcal{S}$ forms the point cloud $\mathcal{P} = \mathcal{S}(\mathcal{C}; \tau).$ 
Consider two point clouds derived from $\mathcal{C}$ with  grid size $\tau_1$ and $ \tau_2$, a robust model $\mathcal{M}$ should satisfy $\mathcal{M}(\mathcal{S}(\mathcal{C}; \tau_1)) \approx \mathcal{M}(\mathcal{S}(\mathcal{C}; \tau_2))$.
However, as shown in \Cref{fig:qualitative}, we observe that existing 3D point cloud encoders~\cite{wu2024point, wu2025sonata, zhang2025concerto} often fail to achieve this semantic consistency. This suggests that models can easily take an easy shortcut and couple semantic features with a specific resolution. 

%%%%%%%%%%%%%%%%%%%%%%%%%%%%%%%%%%%%%%%%%%%%%%%%%%%

%%%%%%%%%%%%%%%%%%%%%%%%%%%%%%%%%%%%%%%%%%%%%%%%%%%

\subsection{What Makes the Models Sensitive to Resolution?}
\label{sec:cause}
 \begin{wrapfigure}[19]{r}{0.4\textwidth} % {r} aligns it to the right
\centering
    \includegraphics[width=0.4\textwidth]{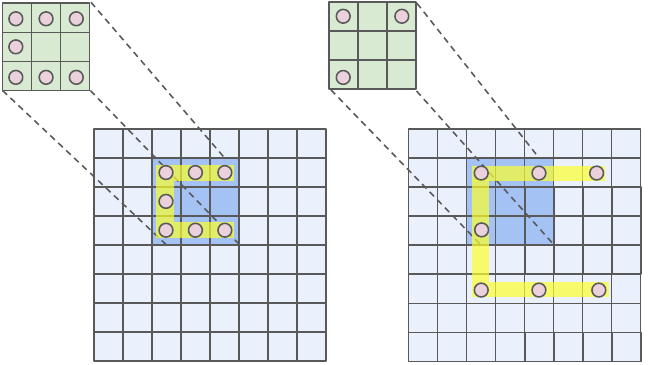}
  \caption{Impact of scale on the relative receptive field of an operator $\Omega$, such as pooling, sparse convolution. While the kernel $\mathcal{K}$ of the operator (green cells) remains a fixed size, the  actual spatial volume it aggregates  change proportionally with the scale. This coupling causes the model to process inconsistent semantic volumes when the scale changes.}
\label{fig:pooling}
\end{wrapfigure}
\paragraph{The Scale Dependency}
The first reason is their inherent scale dependency. At first glance, it may seem difficult to associate resolution with scale in objects; however, this dependency is one of the root causes of the model's sensitivity to sampling resolution. The core issue is architectural: operators such as \textit{pooling}, \textit{relative positional encoding}, and \textit{sparse convolution} are tightly coupled to grid size that define the resolution, which they use to quantize coordinates and thereby define the adjacency relations between features at every stage of the model. Because the grid size determines the scale of the receptive field, altering it is functionally \textbf{equivalent to rescaling every object in the scene} from the model's point of view. 

Moreover, humans recognize an object such as a "chair" through its structural invariants, the arrangement of seat, back, and legs, regardless of its size. This form of scale-invariant recognition is a hallmark of genuine semantic understanding, and it is essential at deployment time, where test-time object sizes frequently diverge from those seen in training or are expressed in entirely different metric units.

We can formalize this bottleneck by viewing these operations as a generalized 
neighborhood aggregation. Let $\mathcal{Q}_{\tau}$ denote a discrete spatial grid with a resolution determined by the grid size $\tau$. For a non empty grid cell $c \in \mathcal{Q}_{\tau}$, 
its output feature $\mathcal{F}_{c}$ is computed as:
\begin{equation}
    \mathcal{F}_{c} = \Omega\left(\left\{f_{i} \mid \frac{d(c, i)}{\tau} \leq \mathcal{K} \right\}\right)
\end{equation}
where $\{f_{i}\}$ is the set of input features, $d(\cdot)$ is a distance metric, $\mathcal{K}$ is a fixed kernel size constant and $\Omega$ denotes a operator that maps the neighborhood set to a single feature. In this formulation, the term $d(c, i)/\tau$ represents the normalized grid distance. As illustrated in \Cref{fig:pooling}, this creates a direct representational dependency: when the resolution $\tau$ or object's size changes, the normalized distance to neighbors changes, effectively altering the volume covered by the fixed kernel $\mathcal{K}$. Consequently, the model perceives a structural change in the object's geometry, providing an easy representational shortcut to learn features that are coupled to a specific scale.

\begin{figure}
    \centering
    \includegraphics[width=1\textwidth]{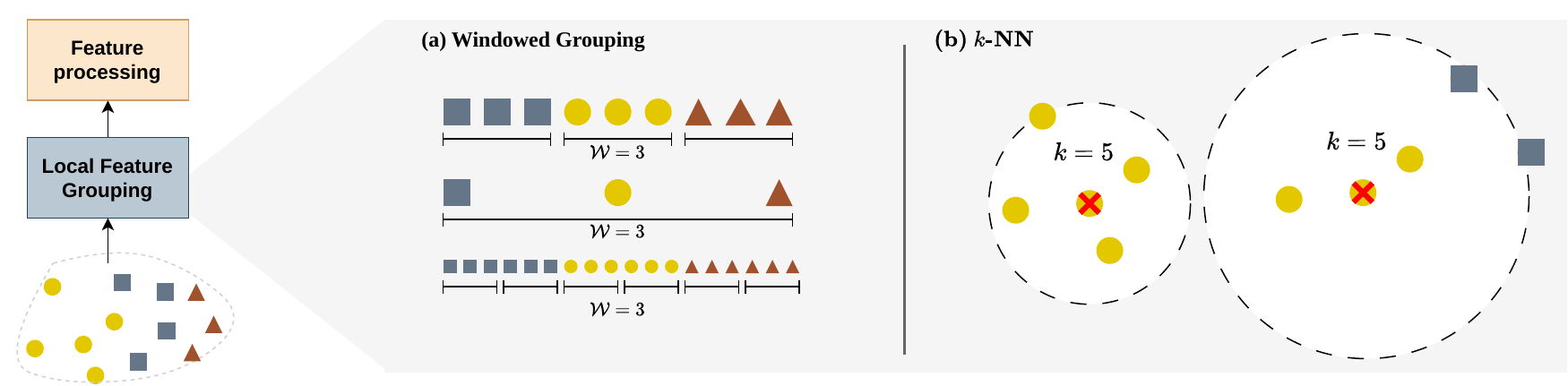}
    \caption{Impact of density on common local feature grouping methods. 
(a) Due to the $O(N^2)$ complexity of global attention, most transformer-based methods adopt patched-attention as a scalable alternative. This introduces a bottleneck: a fixed patch window $\mathcal{W}$ captures inconsistent semantic volumes depending on the sampling resolution.
(b) In $k$-NN based aggregation, the spatial receptive field is inversely proportional to point density. To capture the same $k$ neighbors in a sparse region, the model must aggregate information from a much larger physical area. }
     \label{fig:local_feature_grouping}
     \vspace{-3mm}
\end{figure}
\paragraph{The Density Shortcut}
The other critical factor affecting the performance of point cloud encoders~\cite{choy20194d, wu2024point} is density, defined as the number of points sampled from a scene. Existing works with common local feature grouping methods, such as $k$-NN~\cite{zhao2021point, wu2022point, yang2022unified} and window-attention/voxel-partitioning~\cite{fan2022embracing,park2022fast, sun2022swformer, zhang2022patchformer,liu2023flatformer, yang2025swin3d}, are inherently coupled to point density. 
As shown in \Cref{fig:local_feature_grouping}, as the number of points used to represent an object changes, the local group captures a different spatial context to satisfy the fixed $k$ requirement.
Let $\mathcal{N}_k(x; \mathcal{P})$ denote the set of $k$ neighbors of a query point $x \in \mathcal{P}$. The grouping operation $\mathcal{G}$ then constructs the local point group $g_x$ as:
\begin{equation}
    g_x = \mathcal{G}(\{ x_j \mid x_j \in \mathcal{N}_k(x; \mathcal{P}) \})
\end{equation}

\paragraph{Easy Optimization Paths vs. Invariant Features.} While deep networks are theoretically expressive enough to adapt to these shifts, they typically favor the most efficient optimization path: a representation "shortcut" that couples semantic features to training-specific densities and scales~\cite{geirhos2020shortcut, wu2025sonata}.

\vspace{-3mm}

\section{Achieving Scale and Density Invariance in Point Cloud Encoders}
\label{sec:method}
\begin{figure}
\captionsetup{skip=3pt}
  \centering
   \includegraphics[width=0.95\linewidth]{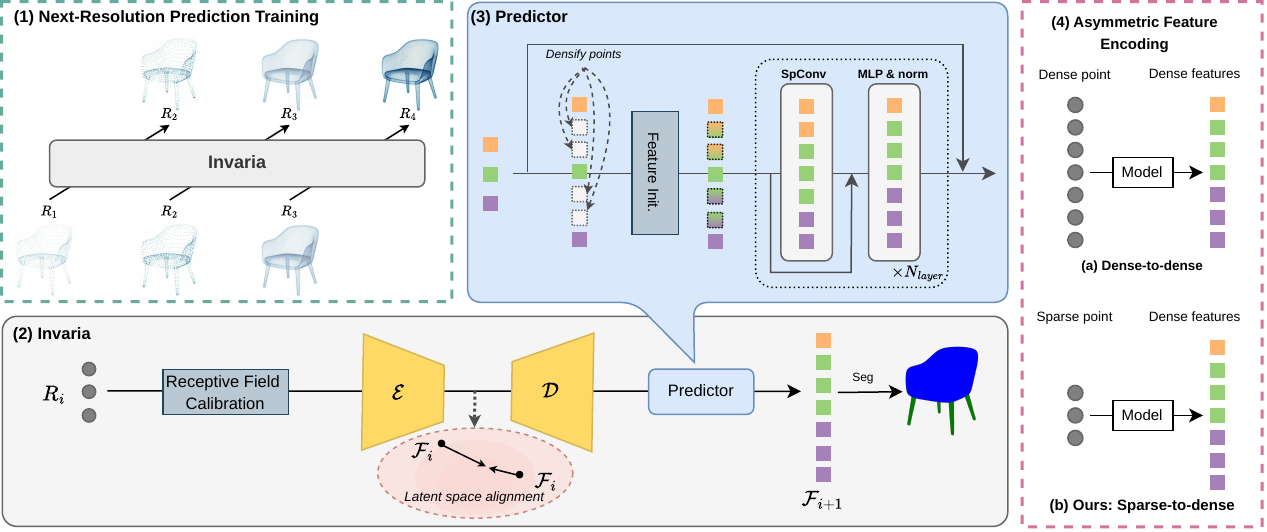}
   \caption{Invaria framework. We employ a next-resolution prediction objective to learn scale- and density-invariant features. This approach decouples  semantic features from the input point cloud’s resolution with a smaller model size. Leveraging these invariant representations, our model employs asymmetric encoding strategy during inference. This allows a large part of the backbone to process  sparser inputs, thereby reducing computation. Unlike conventional point upsampling, which outputs raw points, our approach propagates dense semantic features directly within the latent space.}
   \label{fig:architecture}
   \vspace{-3mm}
\end{figure}

 While existing architectures still inevitably rely on the local operators or grouping methods discussed in \Cref{sec:problem}, we believe that through a proper training paradigm and simple architecture modification, a model can avoid taking the shortcut and instead learn representations that are invariant to both scale and density. Our proposed method is shown in \Cref{fig:architecture}.

\subsection{Next-Resolution Prediction for Resolution Invariant Features Understanding}
\label{subsec:next_resolution_prediction}

In Invaria, we use a next-resolution prediction objective to help the model learn resolution invariant features. 
This approach is inspired by the success of Autoregressive (AR) models such as GPT \cite{brown2020language, radford2018improving} and their recent advancements in the visual domain \cite{tian2024visual, han2025infinity, liu2025infinitystar}. 
Unlike generative models designed for explicit point cloud generation or upsampling, our method utilizes the prediction of a higher resolution point cloud as an objective, forcing the model to learn high-level features across varying resolutions, preventing it from taking the shortcut as discussed in \Cref{sec:problem}.
\paragraph{Receptive Field Calibration.} 
To mitigate the relative receptive field shift inherent in operator $\Omega$ (illustrated in \Cref{fig:pooling}), we introduce a receptive field calibration approach. Rather than relying on a static grid size to define feature adjacency, we define the internal grid dynamically based on the given point cloud $\mathcal{P} = \{x_j\}_{j=1}^N$, where $N$ denotes any number of points sampled to simulate a lower resolution input.
First, we draw a small set of $N_a$ anchor points $\mathcal{P}_a \subset \mathcal{P}$ uniformly at random ($N_a \ll N$), and for each anchor $a_i \in \mathcal{P}_a$ we compute its nearest-neighbor distance to its closest distinct neighbor in $\mathcal{P}$ as $d_i = \min_{x \in \mathcal{P} \setminus \{a_i\}} \|a_i - x\|$.               
The calibrated internal grid size is then obtained by $\psi(\cdot)$, which aggregates these per-anchor distances and rescaling by a constant scaling factor $\alpha$:
\begin{equation}                             
  \psi(\mathcal{P}) = \alpha \cdot \rho\left( \{d_i\}_{i=1}^{N_a} \right)
  \label{eq:receptive_cal}
  \end{equation}  
where $\rho(\cdot)$ is a reduction function, see Appendix \Cref{sec:model_details} for the details of $\rho(\cdot)$. 
Unlike existing work~\cite{zhang2026utonia}, which attempts to bridge this gap using Granularity-Prompted Rescaling, our approach presents two advantages. 
First, we do not require prior knowledge of the physical scale to manually normalize the scene to a predefined canonical granularity. 
Second, by forcing all observations into a single predefined standard scale, such methods fail to capture scale-invariant semantic features and remain sensitive to scenes in which certain objects have a size that differs from those seen during training. 

\paragraph{Point Features Extraction.}
Let $\mathcal{P} = \{x_j\}_{j=1}^{N}$ 
represent a given point cloud, and let the input sparser point 
cloud $\mathcal{P}_{i}$ be a sampled subset such that $\mathcal{P}_{i} \subset \mathcal{P}$ with 
cardinality $|\mathcal{P}_{i}| = N_i$, where $N_i < N$. In this framework, a U-Net-like backbone~\cite{ronneberger2015u} consisting 
of an encoder $\mathcal{E}$ and a decoder $\mathcal{D}$ extracts latent features from the sparse input:
\begin{equation}
    \mathcal{F}_i = \mathcal{D}(\mathcal{E}(\mathcal{P}_{i}))
    \label{eq:fi}
\end{equation}
where $\mathcal{F}_i \in \mathbb{R}^{N_i \times C}$ represents the latent feature representation and $C$ denotes the latent feature dimension. Our implementation of $\mathcal{E}$ and $\mathcal{D}$ follows the standard U-Net architecture.

\paragraph{Latent Space Alignment.}
Receptive field calibration adapts the internal grid to local density, but the backbone is still invoked at different input resolutions. To prevent the model from learning distinct latent spaces for different densities, we align bottleneck features across resolutions. For an input $\mathcal{P}$ presented at $M$ resolutions $\mathcal{P}^{(1)}, \dots, \mathcal{P}^{(M)}$, we obtain a set of bottleneck tokens $\mathcal{Z}^{(m)}$ for each resolution, where each token $z \in \mathcal{Z}^{(m)}$ is associated with a spatial location $c(z) \in \mathbb{R}^3$.
For every pair of resolutions $(m, l)$ with $m < l$, we define the spatial nearest neighbor of $z \in \mathcal{Z}^{(m)}$ as $\pi_l(z) = \arg\min_{z' \in \mathcal{Z}^{(l)}} \lVert c(z) - c(z') \rVert$. We minimize the cosine distance between matched pairs:
\begin{equation}
    \mathcal{L}_{\text{align}} = \frac{1}{\binom{M}{2}} \sum_{m=1}^{M-1} \sum_{l=m+1}^{M} \frac{1}{|\mathcal{Z}^{(m)}|} \sum_{z \in \mathcal{Z}^{(m)}} \Bigl(1 - \cos\bigl(z, \pi_l(z)\bigr)\Bigr)
    \label{eq:align}
\end{equation}

\paragraph{The Predictor.}
To synthesize higher-resolution features from the latent space, we utilize a predictor module $\Phi$. 
Given the latent features $\mathcal{F}_i$ extracted from the sparse input $\mathcal{P}_i$ (as shown in \Cref{eq:fi}), we densify the point set by adding intermediate points. 
We determine the spatial locations for these new points using ground-truth higher-resolution coordinates when accessible, or otherwise via interpolation. 
In either case, the corresponding features are then initialized via nearest-neighbor interpolation from the processed latent features to bridge the sampling density gap.
These features are then iteratively processed through $N_{layer}$ refinement blocks (each consisting of sparse convolution, MLP and norm operations) to produce the next higher-resolution  features, where $\{\hat{f}_j\}_{j=1}^{N_{i+1}}$ denotes the set of reconstructed higher resolution point features.
\begin{equation}
   \mathcal{F}_{i+1} = \Phi(\mathcal{F}_i) = \{\hat{f}_j\}_{j=1}^{N_{i+1}}
\end{equation}

\begin{wrapfigure}[19]{r}{0.4\textwidth} % {r} aligns it to the right
\centering
    \includegraphics[width=0.37\textwidth]{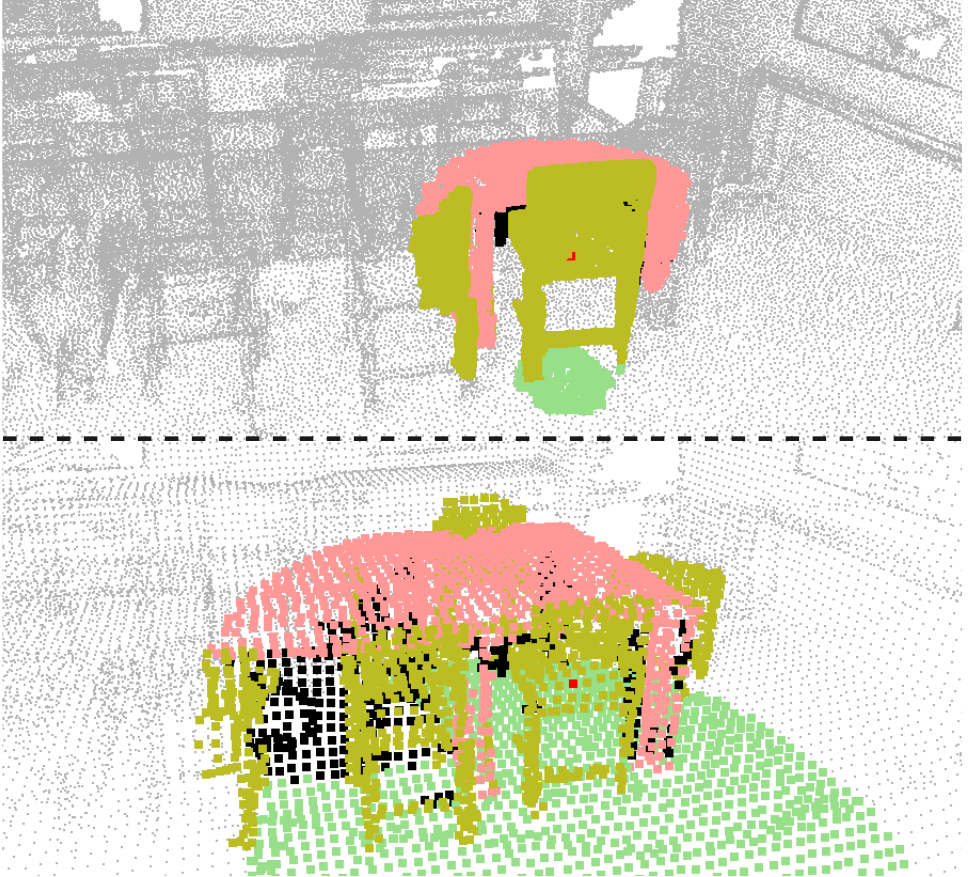}
  \caption{Under the same configuration, the higher-resolution input (top) covers a  smaller receptive field. }
\label{fig:receptive}
\end{wrapfigure}
\vspace{-2mm}
\subsection{Asymmetric Point Feature Encoding.}
\paragraph{The High Resolution Trade-off.} 
It is intuitive that a lower-resolution point cloud degrades performance due to the loss of fine-grained information. Surprisingly, higher resolutions do not necessarily improve performance; in contrast, they may degrade accuracy more than lower-resolution inputs due to a reduced receptive field. As shown in \Cref{fig:receptive}, for an operation that aggregates a fixed number of $k$ input points, the physical receptive field shrinks significantly as the sampling density increases. For an operation taking $k$ points at the grid size $\tau$, the reach $\mathcal{R}$ scales as:
\begin{equation}
    \mathcal{R} \propto \tau \cdot \sqrt[3]{k}
    \label{eq:reach}
\end{equation}
This indicates that at higher resolutions, the spatial receptive field for a fixed $k$ is limited, hindering the model's ability to capture long-range structural dependencies. Therefore, we utilize an asymmetric point feature encoding strategy by processing a sparser input. 

\paragraph{Asymmetric Inference Strategy}
Having trained Invaria to learn resolution-invariant representations through the next-resolution prediction training described in \Cref{subsec:next_resolution_prediction}, we can capitalize on this robustness during the inference stage by adopting an asymmetric encoding strategy that uses sparser point input to produce higher resolution features. By utilizing sparser, lower-resolution inputs for the U-Net backbone, the model maintains a larger effective receptive field and simultaneously reduces the computational FLOPs required compared to dense processing. See \Cref{sec:model_details} for the lower-resolution sampling setting.

\subsection{Training Objective}
\label{subsec:training_objective}
\vspace{-3mm}
To ensure robust semantic understanding across varying scales, we supervise the network across $M$ resolution levels. The model takes a point cloud at resolution $m$ as input and is trained to predict the features of the subsequent resolution $m+1$. The total training objective combines a supervised segmentation loss, comprising Cross-Entropy (CE) and Lovász-Softmax (LS)~\cite{berman2018lovasz}, with a consistency alignment term, $\mathcal{L}_{\text{align}}$ formulated in \Cref{eq:align}, where $\lambda^{(m)}$ denotes the weighting factor for the $m$-th resolution level, and $\beta$ is a scaling constant.
\begin{equation}
\mathcal{L} = \sum_{m=1}^{M-1} \lambda^{(m)} \left( \mathcal{L}_{\text{CE}} + \mathcal{L}_{\text{LS}} \right) + \beta \mathcal{L}_{\text{align}}
\label{eq:loss}
\end{equation}
\definecolor{lightgray}{gray}{0.9}
\newcommand{\cmark}{\ding{51}}%

\begin{table}
\centering
\caption{Robustness to resolution shifts on ScanNet (val). We evaluate models trained at a default resolution ($\tau^*$=2cm grid size) across varying resolutions. \textit{SSL:} Self-Supervised Learning; \textit{MD:} Joint Training across multiple datasets. $^\dagger$Results using authors' publicly released weights (\textit{linear probing}).}
\label{tab:resolution_change}
\footnotesize % Standard size for dense tables; much more readable than resizebox
\setlength{\tabcolsep}{3.0pt} % Tightens internal padding to fit the page
\begin{tabular}{l ccc ccc c ccc c ccc}
\toprule
 \multirow{2}{*}{\textbf{Model}} & \multirow{2}{*}{\textbf{Params}} & \multirow{2}{*}{\textbf{\shortstack{SSL\& \\ MD}}} & \multicolumn{4}{c}{\textbf{mIoU (\%)}} & \multicolumn{4}{c}{\textbf{mAcc (\%)}} & \multicolumn{4}{c}{\textbf{allAcc (\%)}} \\
\cmidrule(lr){4-7} \cmidrule(lr){8-11} \cmidrule(lr){12-15}
 &  &  & $0.5\tau^*$ & \cellcolor[HTML]{D3D3D3}$\tau^*$ & $2\tau^*$ & $3\tau^*$ & $0.5\tau^*$ & \cellcolor[HTML]{D3D3D3}$\tau^*$ & $2\tau^*$ & $3\tau^*$ & $0.5\tau^*$ & \cellcolor[HTML]{D3D3D3}$\tau^*$ & $2\tau^*$ & $3\tau^*$ \\
\midrule
$\text{Sonata}^\dagger$\cite{wu2025sonata} & 124.8M & \checkmark & 59.9 & 72.6 & 62.3 & 34.2 & 70.7 & 83.4 & 73.8 & 43.7 & 84.9 & 89.8 & 85.7 & 72.3 \\
$\text{Concrt-L}^\dagger$\cite{zhang2025concerto} & 207.7M & \checkmark & 67.9 & \cellcolor[HTML]{def3e4}\bf 78.6 & \cellcolor[HTML]{eef9f0}72.5 & 46.0 & 77.5 & \cellcolor[HTML]{def3e4}\bf 87.4 & \cellcolor[HTML]{eef9f0}82.3 & 57.3 & 88.4 & \cellcolor[HTML]{eef9f0}92.3 & \cellcolor[HTML]{eef9f0}89.9 & 79.1 \\
$\text{Utonia}^\dagger$\cite{zhang2026utonia} & 157.7M & \checkmark & 73.0 & 76.6 & 71.7 & 47.3 & 81.9 & \cellcolor[HTML]{eef9f0}85.1 & 80.5 & 56.3 & \cellcolor[HTML]{def3e4}\bf 95.4 & \cellcolor[HTML]{def3e4}\bf 94.6 & 89.2 & 78.7 \\
\midrule
Sp.Mink.\cite{choy20194d} & 39.2M & - & 36.6 & 75.6 & 47.8 & 19.4 & 47.2 & 83.5 & 56.0 & 25.7 & 73.8 & 91.5 & 82.4 & 66.1 \\
PTv3\cite{wu2024point}  & 46.1M & - & 50.5 & \cellcolor[HTML]{eef9f0}77.6 & 60.4 & 20.5 & 59.3 & 85.0 & 69.0 & 26.0 & 80.6 & 92.0 & 86.0 & 65.9 \\
Invaria-s & 16.3M & - & \cellcolor[HTML]{eef9f0}73.1 & 73.3 & 72.2 & \cellcolor[HTML]{eef9f0}70.7 & \cellcolor[HTML]{eef9f0}80.2 & 80.5 & 79.3 & \cellcolor[HTML]{eef9f0}78.1 & 89.9 & 89.9 & 89.5 & \cellcolor[HTML]{eef9f0}88.9 \\
Invaria & 25.5M & - & \cellcolor[HTML]{def3e4}\bf 75.5 & 76.2 & \cellcolor[HTML]{def3e4}\bf 75.7 & \cellcolor[HTML]{def3e4}\bf 73.8 & \cellcolor[HTML]{def3e4}\bf 83.8 & 84.3 & \cellcolor[HTML]{def3e4}\bf 84.0 & \cellcolor[HTML]{def3e4}\bf 82.7  & \cellcolor[HTML]{eef9f0}91.0 & 91.2 & \cellcolor[HTML]{def3e4}\bf 90.9 & \cellcolor[HTML]{def3e4}\bf 90.0 \\
\bottomrule
\end{tabular}
\vspace{-5mm}
\end{table}
\vspace{-3mm}
\section{Discussion and Analysis}
\label{sec:experiment}
\begin{wrapfigure}[17]{r}{0.5\textwidth} % {r} aligns it to the right
\centering
    \includegraphics[width=0.45\textwidth]{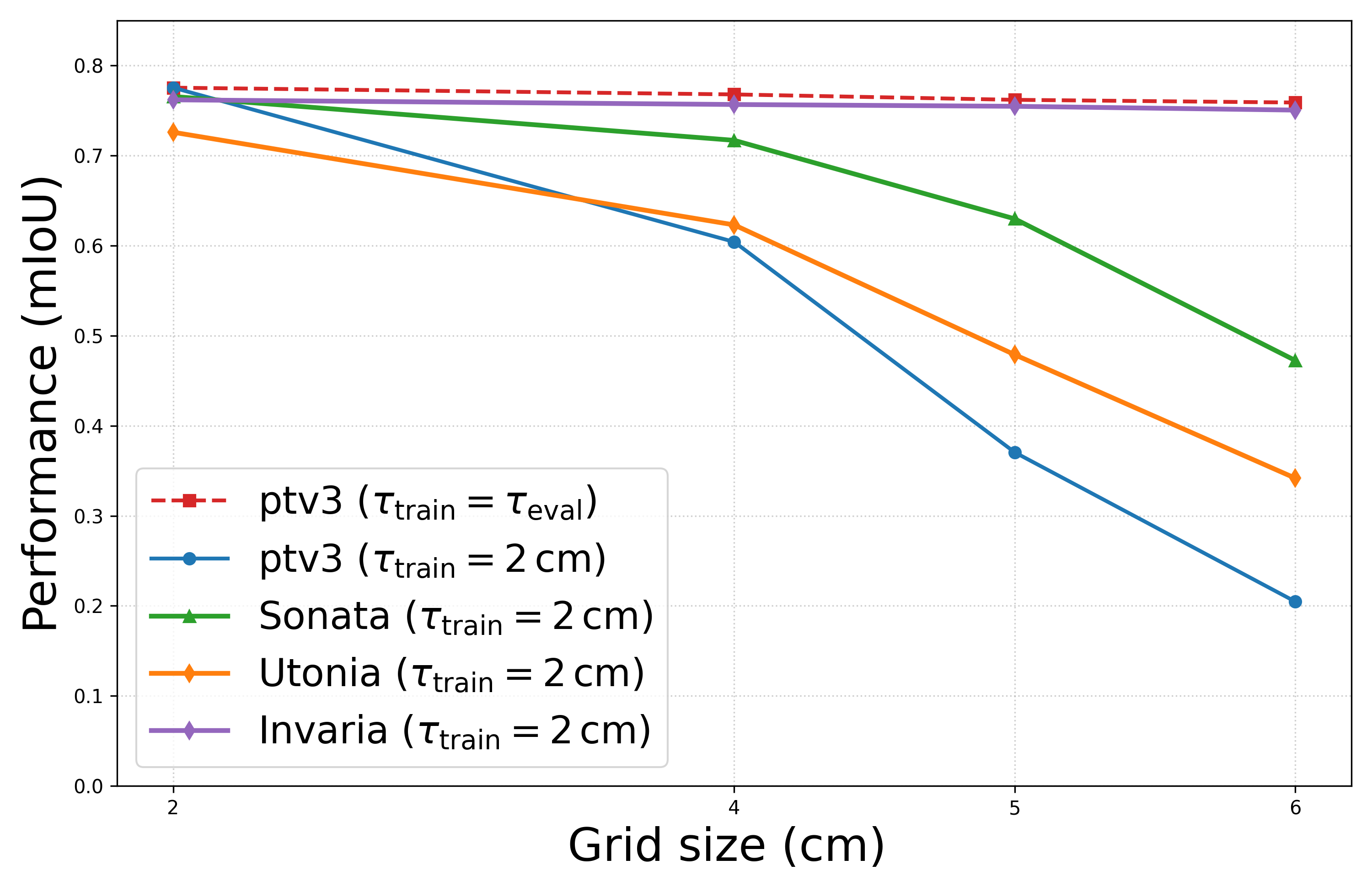}
  \caption{The performance of existing methods when evaluated on different resolutions on ScanNet. The \textcolor{dark_red}{red dashed line} represents a model trained and evaluated at the same resolution, showing that even at $6$cm, the point cloud retains sufficient information for high-accuracy semantic segmentation.}
\label{fig:density_generalization_analysis}  
\end{wrapfigure}

\vspace{-3mm}
\paragraph{Does performance degradation stem from insufficient geometric cues?} 
No. As shown in \Cref{fig:density_generalization_analysis}, training PTv3~\cite{wu2024point} from scratch across varying resolutions, and evaluating each on its respective training distribution ($\tau_{train}=\tau_{eval}$), reveals that low resolution point clouds retain sufficient information for accurate recognition. Instead, models take the shortcut and learn resolution-specific features that are coupled with resolution. Moreover, \Cref{tab:resolution_change}, \Cref{tab:nuscenes_val} shows that performance degrades even more when higher-resolution ($\tau=1$cm) input is provided. We also implemented a lightweight version, \textit{Invaria-s}; please refer to Appendix~\Cref{sec:model_details} for further details.

\paragraph{Can Self-Supervised Training and Scaling Overcome Resolution-Specific Bias?}
Partially. Self-supervised learning (SSL)~\cite{pang2023masked, he2022masked} and the application of scaling laws~\cite{brown2020language} have emerged as powerful drivers for enhancing the generalizability of 3D encoders~\cite{wu2025sonata, zhang2025concerto, zhang2026utonia}. By pre-training on massive, unlabeled datasets, models can extract semantically rich point features. However, as shown in \Cref{fig:density_generalization_analysis}, \Cref{tab:resolution_change}, while these methods offer improved robustness to resolution shifts compared to other baselines, this progress remains limited. Crucially, these approaches rely on massive training datasets, prolonged training times, and increased model size. This overhead hinders deployment in real-world robotic environments, where latency and hardware constraints are critical.

\definecolor{lightgray}{gray}{0.9}
\begin{table}[t]
\centering
\footnotesize % Give it a starting point so resizebox doesn't work as hard
\begin{minipage}{0.5\textwidth}
    \centering
    \caption{Robustness to resolution shifts on outdoor dataset (nuScenes val) ($\tau^*$=5cm). $^\dagger$ \textit{linear probing}. }
    \label{tab:nuscenes_val}
    \resizebox{\textwidth}{!}{
        \setlength{\tabcolsep}{1.5pt} % CRITICAL: Tighten spacing
        \definecolor{lightgray}{gray}{0.9}

\begin{tabular}{l c c ccc c ccc}
    \toprule
     \multirow{2}{*}{\textbf{Model}} & \multirow{2}{*}{\textbf{Params}} & \multicolumn{4}{c}{\textbf{mIoU (\%)}} & \multicolumn{4}{c}{\textbf{mAcc (\%)}} \\
    \cmidrule(lr){3-6} \cmidrule(lr){7-10}
     & & $0.5\tau^*$ & \cellcolor[HTML]{D3D3D3}$\tau^*$ & $2\tau^*$ & $3\tau^*$ & $0.5\tau^*$ & \cellcolor[HTML]{D3D3D3}$\tau^*$ & $2\tau^*$ & $3\tau^*$ \\\midrule
     Sonata$^\dagger$\cite{wu2025sonata}     & 124.8M & - & 66.1 & - & - & - & 77.2 & - & - \\
     Concrt$^\dagger$\cite{zhang2025concerto}     & 207.7M & - & 74.2 & - & - & - & 83.2 & - & - \\
    Utonia$^\dagger$\cite{zhang2026utonia}      & 157.7M & - & 75.5 & - & - & - & \cellcolor[HTML]{eef9f0}84.8 & - & - \\
    \midrule
    Sp.Mink.\cite{choy20194d}    & 39.2M & 34.1 & 76.7 & 35.4 & 14.5 & 41.4 & 82.0 & 45.7 & 22.6 \\
    PTv3\cite{wu2024point}        & 46.1M & 45.1 & \cellcolor[HTML]{def3e4}\bf 80.4 & 51.9 & 26.4 & 52.9 & \cellcolor[HTML]{def3e4}\bf 86.2 & 59.7 & 33.0 \\
    Invaria-s   & 16.3M & \cellcolor[HTML]{eef9f0}72.9 & 73.0 & \cellcolor[HTML]{eef9f0}71.8 & \cellcolor[HTML]{eef9f0}68.6 & \cellcolor[HTML]{eef9f0}80.1 & 80.2 & \cellcolor[HTML]{eef9f0}79.8 & \cellcolor[HTML]{eef9f0}77.4 \\
    Invaria     & 25.5M & \cellcolor[HTML]{def3e4}\bf 74.3 & \cellcolor[HTML]{eef9f0}76.2 & \cellcolor[HTML]{def3e4}\bf 75.6 & \cellcolor[HTML]{def3e4}\bf 71.9 & \cellcolor[HTML]{def3e4}\bf 82.3 & \cellcolor[HTML]{eef9f0} 83.9 & \cellcolor[HTML]{def3e4}\bf 83.4 & \cellcolor[HTML]{def3e4}\bf 80.1 \\
    \bottomrule
\end{tabular}
    }
\end{minipage}
\hfill
\begin{minipage}{0.45\textwidth}
    \centering
    \caption{Evaluation on model size and efficiency of transformer-based methods. $^\dagger$ \textit{linear probing}.}
    \label{tab:efficiency}
    \resizebox{\textwidth}{!}{
        \setlength{\tabcolsep}{1.5pt}
        
\scriptsize
\begin{tabular}{l ccc}
\toprule
\textbf{\scriptsize Method} & \bf \scriptsize Params (M) &  \bf \scriptsize FLOPs (G) & \bf \scriptsize Mem. (GB) \\ 
\midrule
Sonata$^\dagger$\cite{wu2025sonata}     & 124.8 & 691 & 1.65 \\
Concrt-L$^\dagger$\cite{zhang2025concerto}    & 207.7 & 1248 & 2.49 \\
Utonia$^\dagger$\cite{zhang2026utonia}      & 157.7 & 874 & 1.9 \\
PTv3\cite{wu2024point}        & 46.1  & 352 & 0.59 \\
Invaria-s   &  16.3  &  230 &  0.31 \\
Invaria     & 25.5  & 267 & 0.54 \\ 
\bottomrule
\end{tabular}
    }
\end{minipage}
\vspace{-5mm}
\end{table}
\definecolor{lightgray}{gray}{0.9}
\begin{table}[t]
\centering
\footnotesize % Give it a starting point so resizebox doesn't work as hard
\begin{minipage}{0.68\textwidth}
    \centering
    \caption{Analysis on each category's performance on ScanNet (val). We report the relative performance drop (\%) on IoU from the default training resolution to $3\times $ lower. The results suggest an even more severe performance drop for complex objects. $^\dagger$ \textit{linear probing}.}
    \label{tab:per_class_robustness}
    \resizebox{\textwidth}{!}{
        \setlength{\tabcolsep}{1.5pt}
        \definecolor{SoftRed}{HTML}{F2D7D5}

\begin{tabular}{ l | *{9}{c}}
\toprule
\textbf{Method} & Wall & Floor & Sofa & Chair & Toilet & Fridge & Counter & Picture & Sink \\ 
\midrule
% --- IoU GROUP ---
Sonata$^\dagger$\cite{wu2025sonata} & \cellcolor{Salmon!20} -16.6\% & \cellcolor{Salmon!6} -6.8\% & \cellcolor{Salmon!70} -74.7\% & \cellcolor{Salmon!52} -52.1\% & \cellcolor{Salmon!90} -90.5\% & \cellcolor{Salmon!72} -72.2\% & \cellcolor{Salmon!87} -87.6\% & \cellcolor{Salmon!71} -71.0\% & \cellcolor{Salmon!78} -78.0\% \\
Concrt-L$^\dagger$\cite{zhang2025concerto} & \cellcolor{Salmon!13}-13.0\% & \cellcolor{Salmon!6}-5.8\% & \cellcolor{Salmon!16}-16.4\% & \cellcolor{Salmon!35}-34.5\% & \cellcolor{Salmon!96}-96.2\% & \cellcolor{Salmon!67}-67.1\% & \cellcolor{Salmon!61}-61.3\% & \cellcolor{Salmon!37}-36.6\% & \cellcolor{Salmon!72}-72.1\% \\
Utonia$^\dagger$\cite{zhang2026utonia} 
& \cellcolor{Salmon!13} -13.2\% & \cellcolor{Salmon!5} -5.7\% & \cellcolor{Salmon!10} -10.6\% & \cellcolor{Salmon!30} -30.1\% & \cellcolor{Salmon!70} -71.8\% & \cellcolor{Salmon!60} -60.8\% & \cellcolor{Salmon!43} -43.0\% & \cellcolor{Salmon!61} -61.9\% & \cellcolor{Salmon!44} -44.8\% \\
\cmidrule{1-10}
Sp.Mink.\cite{choy20194d} 
& \cellcolor{Salmon!25} -25.7\% & \cellcolor{Salmon!11} -11.3\% & \cellcolor{Salmon!89} -89.1\% & \cellcolor{Salmon!60} -61.6\% & \cellcolor{Salmon!100} -100\% & \cellcolor{Salmon!100} -100\% & \cellcolor{Salmon!100} -100\% & \cellcolor{Salmon!98} -98.4\% & \cellcolor{Salmon!99} -99.4\% \\
PTv3\cite{wu2024point}
& \cellcolor{Salmon!31} -31.0\% & \cellcolor{Salmon!22} -22.2\% & \cellcolor{Salmon!95} -95.1\% & \cellcolor{Salmon!68} -68.1\% & \cellcolor{Salmon!100} -100\% & \cellcolor{Salmon!99} -99.9\% & \cellcolor{Salmon!99} -99.9\% & \cellcolor{Salmon!85} -85.1\% & \cellcolor{Salmon!93} -93.6\% \\
Invaria-s & \cellcolor{Salmon!2} \bf-2.1\% & \cellcolor{Salmon!1} \bf-0.5\% & \cellcolor{Salmon!1}\bf-0.6\% & \cellcolor{Salmon!2} \bf-1.6\% & \cellcolor{Salmon!3}\bf-2.8\% & \cellcolor{Salmon!9}-8.7\% & \cellcolor{Salmon!4} -3.8\% & \cellcolor{Salmon!17} \bf-17.1\% & \cellcolor{Salmon!4} -4.2\% \\
Invaria
& \cellcolor{Salmon!2} -2.3\% & \cellcolor{Salmon!1} -0.6\% & \cellcolor{Salmon!2}-2.1\% & \cellcolor{Salmon!2} -1.9\% & \cellcolor{Salmon!3}-3.3\% & \cellcolor{Salmon!6}\bf-5.8\% & \cellcolor{Salmon!2} \bf-1.5\% & \cellcolor{Salmon!28} -28.3\% & \cellcolor{Salmon!0} \bf-0.3\% \\
\bottomrule
\end{tabular}
    }
\end{minipage}
\hfill
\begin{minipage}{0.3\textwidth}
    \centering
    \caption{Scale change, where the scenes are scaled up or down. $^\dagger$ \textit{linear probing}.}
    \label{tab:scale_change}
    \resizebox{\textwidth}{!}{
        \setlength{\tabcolsep}{1.5pt} % CRITICAL: Tighten spacing

\begin{tabular}{l c ccc}
\toprule
 \multirow{2}{*}{\textbf{Model}} & \multicolumn{4}{c}{\textbf{mIoU (\%)}}  \\
\cmidrule{2-5} 
 & $\times 2$ & \cellcolor[HTML]{D3D3D3}$\times 1$ & $\times 1/2$ & $\times 1/3$ \\
\midrule
Sonata$^\dagger$\cite{wu2025sonata}      & 61.6 & 72.6 & 66.9 & 44.8  \\

Concrt-L$^\dagger$\cite{zhang2025concerto} & 68.2 & \cellcolor[HTML]{def3e4}\bf 78.6 & \cellcolor[HTML]{eef9f0}75.4 & 61.4  \\

Utonia$^\dagger$\cite{zhang2026utonia}      & \cellcolor[HTML]{eef9f0}74.4 & \cellcolor[HTML]{eef9f0}76.6 & 69.7 & 42.2  \\
\midrule
Sp.Mink.\cite{choy20194d}      & 33.4 & 75.6 & 42.9 & 14.7 \\
PTv3\cite{wu2024point}        & 47.2 & 77.6 & 23.2 & 9.4 \\
Invaria-s & 73.9 & 73.3 & 74.0 & \cellcolor[HTML]{eef9f0}73.9  \\
Invaria & \cellcolor[HTML]{def3e4}\bf 75.6 & 76.2 & \cellcolor[HTML]{def3e4}\bf 75.6 & \cellcolor[HTML]{def3e4}\bf 73.8 \\

\bottomrule
\end{tabular}

    }
\end{minipage}
\vspace{-5mm}
\end{table}

\paragraph{Model Size and Computational Overhead.} 
As shown in \Cref{tab:efficiency}, our method achieves superior performance compared to the existing works while utilizing about 45\% fewer parameters and lower FLOPs. Moreover, our asymmetric feature encoding strategy allows us to reduce the input token count for our U-Net backbones by approximately 40\% on average, directly contributing to the overall efficiency of the framework.

\paragraph{Does Resolution Augmentation Resolve Sensitivity?}
Limited. By testing on standard PTv3~\cite{wu2024point} we noticed the following limitations: First, mixing resolutions induces feature interference, resulting in a ~2\% mIoU drop (77.6\% $\rightarrow$ 75.6\%) on ScanNet~\cite{dai2017scannet} at the native training resolution ($\tau^*$), which suggests conflicting feature representations. Moreover, this improvement seems to be interpolative; when evaluated on denser inputs ($0.5\tau^*$) that fall outside the augmented training range, performance collapses (56.0\% vs. our 75.5\%). Second, this strategy is inherently unidirectional: while we can easily simulate low-resolution inputs via downsampling, we cannot easily synthesize higher-resolution features that are absent in the original dataset.

\paragraph{What kind of object classes suffer the most?}
While aggregate metrics such as mIoU provide a high-level overview, a per-class analysis reveals significant performance disparities across categories (\Cref{tab:per_class_robustness}). Large, structurally simple categories such as \textit{Wall} and \textit{Floor} exhibit high resilience across all methods, as their planar geometry remains identifiable even under $3\times$ lower resolution. However, deeper investigation into the most affected classes reveals a severe performance collapse in two distinct groups. First, complex geometries such as \textit{Toilet}, \textit{Sofa}, and \textit{Sink} suffer near-total failure in baselines such as PTv3~\cite{wu2024point} and Sp.Mink.~\cite{choy20194d}, with accuracy drops reaching $90\%$ to $100\%$. Second, surprisingly simple planar objects, such as \textit{Picture} and \textit{Fridge}, also demonstrate significant degradation. Our analysis suggests this stems from geometric ambiguity: as resolution or scale shifts, these objects lose their distinguishing features and are easily misclassified as \textit{Wall}.

\paragraph{Scale and Density Analysis.} 
We evaluate model robustness to scale changes (while maintaining the original number of points). As shown in \Cref{tab:scale_change}, existing methods are remarkably brittle to scale variations; even models trained with SSL approaches on larger datasets cannot mitigate this issue. Furthermore, Appendix \Cref{tab:density_change} examines density shifts (where the scale remains constant but the number of points is reduced), demonstrating how sparser point cloud representations lead to performance degradation.

\begin{table}[t]
\centering
\caption{Robustness of different model families to resolution shifts on ScanNet (val), reported in mIoU (\%). \textit{Retention} is the mIoU at $3\tau^*$ relative to the mIoU at the default resolution $\tau^*$.}
\label{tab:model_family}
\footnotesize
\setlength{\tabcolsep}{3.0pt}
\begin{tabular}{l l l cccc c}
\toprule
\textbf{Model} & \textbf{Family} & \textbf{Neighborhood} & $0.5\tau^*$ & \cellcolor[HTML]{D3D3D3}$\tau^*$ & $2\tau^*$ & $3\tau^*$ & \textbf{Retention} \\
\midrule
PointNet\cite{qi2017pointnet} & Point-based & global max-pooling & - & 36.7 & 35.8 & 35.1 & \cellcolor[HTML]{eef9f0}95.8\% \\
PointNet++\cite{qi2017pointnet++} & Point-based & fixed-radius ball & - & 50.5 & 48.7 & 40.6 & 80.5\% \\
KPConv\cite{thomas2019kpconv} & Kernel-based & radius tied to grid & - & 69.7 & 54.8 & 31.5 & 45.3\% \\
Sp.Mink.\cite{choy20194d} & Sparse voxel conv. & grid kernel & 36.6 & 75.6 & 47.8 & 19.4 & 25.7\% \\
PTv2\cite{wu2022point} & Transformer & $k$-NN & 55.4 & 75.4 & 56.0 & 34.1 & 45.2\% \\
PTv3\cite{wu2024point} & Transformer & serialized patch & 50.5 & \cellcolor[HTML]{def3e4}\bf 77.6 & 60.4 & 20.5 & 26.4\% \\
\midrule
Invaria & Transformer & serialized patch & \cellcolor[HTML]{def3e4}\bf 75.5 & \cellcolor[HTML]{eef9f0}76.2 & \cellcolor[HTML]{def3e4}\bf 75.7 & \cellcolor[HTML]{def3e4}\bf 73.8 & \cellcolor[HTML]{def3e4}\bf 96.8\% \\
\bottomrule
\end{tabular}
\vspace{-3mm}
\end{table}

\paragraph{Are Other Model Families More Robust?}
Beyond transformer-based encoders, we evaluate representatives of other model families under the same resolution shifts (\Cref{tab:model_family}). A natural alternative to $k$-NN grouping (\Cref{fig:local_feature_grouping}) is a fixed metric radius, as used in the set-abstraction layers of PointNet++~\cite{qi2017pointnet++}. Although the physical receptive field then stays constant, PointNet++ still degrades under resolution shifts: the radius remains the same, but the number of points falling inside it changes, and so do the features built from them. KPConv~\cite{thomas2019kpconv}, whose kernel radius is tied to the grid size, retains only 45.3\% of its performance, consistent with the scale dependency discussed in \Cref{sec:cause}. In contrast, PointNet~\cite{qi2017pointnet}, which aggregates features through global max-pooling instead of a grid-defined local neighborhood, is remarkably robust (95.8\% retention), although its accuracy at the default resolution is far lower.

However, high retention can overstate robustness when the accuracy at the default resolution is low, since such models obtain most of their mIoU from large, geometrically simple classes that are inherently robust to resolution changes. For example, PointNet++ retains 98.1\% and 92.4\% of its IoU on \textit{Floor} and \textit{Wall} at $3\tau^*$, but only 46.9\% and 48.1\% on \textit{Shower Curtain} and \textit{Door}.

\begin{table}[t]
\centering
\caption{Robustness to different sampling distributions on ScanNet (val), reported in mIoU (\%). Each sampling scheme keeps approximately the same number of points as voxel subsampling at $3\tau^*$ 
.}
\label{tab:sampling_distribution}
\footnotesize
\setlength{\tabcolsep}{4.0pt}
\begin{tabular}{l c ccc}
\toprule
\textbf{Model} & \cellcolor[HTML]{D3D3D3}\textbf{Default} & \textbf{Uniform random} & \textbf{FPS} & \textbf{Locally varying} \\
\midrule
Sp.Mink.\cite{choy20194d} & 75.6 & 15.7 & 18.7 & 15.9 \\
PTv3\cite{wu2024point} & \cellcolor[HTML]{def3e4}\bf 77.6 & 15.4 & 19.4 & 15.3 \\
Invaria & 76.2 & \cellcolor[HTML]{def3e4}\bf 71.2 & \cellcolor[HTML]{def3e4}\bf 72.6 & \cellcolor[HTML]{def3e4}\bf 71.0 \\
\bottomrule
\end{tabular}
\vspace{-5mm}
\end{table}

\paragraph{Robustness to Sampling Distribution.}
A real-world density change is not necessarily uniform. We therefore evaluate three sampling schemes that keep the same number of points as voxel subsampling at $3\tau^*$ (\Cref{tab:sampling_distribution}). \textit{Uniform random} sampling keeps every point with the same probability $p = 0.115$, independent of its position, so the existing density variation of the scene is preserved and simply thinned. \textit{Farthest point sampling} (FPS) iteratively selects the point at maximum distance from those already chosen, driving the spacing toward uniform. \textit{Locally varying} sampling modulates $p$ smoothly across the scene between 0.05 and 0.19 with the same mean, resulting in a $3.4\times$ density ratio within a single scene that simulates locally sparse and dense regions. Invaria remains robust under all three schemes, whereas the baselines degrade severely, showing that its robustness extends beyond global density reduction to changes in the sampling distribution and strong local density variation.

\begin{table}[t]
    \caption{Ablation studies on ScanNet~\cite{dai2017scannet} val. Due to space constraints, results for (c), (f), (g), and (h) are reported using the default grid size $\tau^*=2$cm. The final selections are indicated by \colorbox[HTML]{cce3de}{green}.}
    \label{tab:ablation}
    \begin{minipage}{0.32\textwidth}
    \centering
        \tablestyle{1pt}{1.08}
        \begin{tabular}{x{10mm}|x{8mm}x{8mm}x{8mm}x{8mm}}
$\mathcal{G}(\cdot)$ & $0.5\tau^* $ & $\tau^*$ & $2\tau^*$ & $3\tau^*$ \\\midrule
\cellcolor[HTML]{cce3de}w & \cellcolor[HTML]{def3e4}\bf 75.5 & \cellcolor[HTML]{def3e4}\bf 76.2 & \cellcolor[HTML]{def3e4}\bf 75.7 & \cellcolor[HTML]{def3e4}\bf 73.8  \\
w/o & 55.7 & 76.0 & 74.5 & 69.7 \\
\end{tabular}
        \vspace{-1mm}
        \subcaption{\textbf{Receptive Field Calibration.} As discussed in \Cref{eq:receptive_cal}, without this calibration, the model's performance drops as resolution changes.  (mIoU)}
        \label{subtab:ablation_receptive_field}
    \end{minipage}
    \hspace{0.03\textwidth}
    \begin{minipage}{0.32\textwidth}
    \centering
        \tablestyle{1pt}{1.08}
        \begin{tabular}{x{10mm}|x{8mm}x{8mm}x{8mm}x{8mm}}
NRP & $0.5\tau^* $ & $\tau^*$ & $2\tau^*$ & $3\tau^*$ \\\midrule
\cellcolor[HTML]{cce3de}w & \cellcolor[HTML]{def3e4}\bf 75.5 & 76.2 & \cellcolor[HTML]{def3e4}\bf 75.7 & \cellcolor[HTML]{def3e4}\bf 73.8 \\
w/o & 45.4 & \cellcolor[HTML]{def3e4}\bf 76.5 & 21.7 & 9.5 \\
\end{tabular}

        \vspace{-1mm}
        \subcaption{\textbf{Next Resolution Prediction.} Comparison between the proposed model and a baseline trained from scratch without the next-resolution prediction pretext task.}
        \label{subtab:ablation_nrp}
    \end{minipage}
        \hspace{0.03\textwidth}
    \begin{minipage}{0.25\textwidth}
    \centering
        \tablestyle{1pt}{1.08}
        \begin{tabular}{x{15mm}|x{10mm} x{10mm}}
Focal loss & mIoU &mAcc \\\midrule
w &  75.1 & 82.9 \\
\cellcolor[HTML]{cce3de}w/o &  \cellcolor[HTML]{def3e4}\bf 76.2 & \cellcolor[HTML]{def3e4}\bf 84.3  \\
\end{tabular}

        \vspace{-1mm}
        \subcaption{\textbf{Focal Loss.} To address category imbalance, we experimented with Focal Loss; however, it does not improve the performance.}
        \label{subtab:ablation_focal_loss}
    \end{minipage} \\
    \begin{minipage}{0.42\textwidth}
    \centering
        \tablestyle{1pt}{0.9}
        \begin{tabular}{x{15mm}|x{8mm}x{8mm}x{8mm}x{10mm}}
$\lambda^{(m)}$ & $0.5\tau^* $ & $\tau^*$ & $2\tau^*$ & $3\tau^*$ \\\midrule
Uniform & 75.2 & 75.9 & 75.7 & \bf \cellcolor[HTML]{def3e4}74.0  \\
\cellcolor[HTML]{cce3de}Increase & \cellcolor[HTML]{def3e4}\bf 75.5 & \cellcolor[HTML]{def3e4}\bf 76.2 & \cellcolor[HTML]{def3e4}\bf 75.7 & 73.8 \\
\end{tabular}

        \vspace{-1mm}
        \subcaption{\textbf{Loss weight.}
        In \Cref{eq:loss} we found that assigning higher weight to higher resolution increases overall performance. (mIoU)
        }
        \label{subtab:ablation_loss_weight}
        \vspace{1mm}
    \end{minipage}
    \hspace{0.06\textwidth}
    \begin{minipage}{0.42\textwidth}
    \centering
        \tablestyle{1pt}{1.13}

\begin{tabular}{x{15mm}|x{10mm}x{10mm}x{10mm}x{10mm}}
$\mathcal{L}_{align}$ & $0.5\tau^* $ & $\tau^*$ & $2\tau^*$ & $3\tau^*$ \\\midrule
\cellcolor[HTML]{cce3de}w & \cellcolor[HTML]{def3e4}\bf 75.5 & \cellcolor[HTML]{def3e4}\bf 76.2 & \cellcolor[HTML]{def3e4}\bf 75.7 & \cellcolor[HTML]{def3e4}\bf 73.8 \\
w/o & 74.9 & 75.4 & 75.0 & 73.3 \\
\end{tabular}

        \vspace{-1mm}
        \subcaption{\textbf{Latent space alignment.} Adding ${\mathcal{L}_{align}}$ helps the model perform across all tested resolution. (mIoU)
        }
        \label{subtab:ablation_latent_space_alignment}
        \vspace{1mm}
    \end{minipage} \\
    \begin{minipage}{0.38\textwidth}
    \centering
        \tablestyle{1pt}{1.08}

\begin{tabular}{x{20mm}|x{10mm}x{10mm}x{10mm}}
\# btlnck layer & 0 & \cellcolor[HTML]{cce3de}1 & 2 \\\midrule
mIoU & 74.3 & \cellcolor[HTML]{def3e4}\bf 76.2 & 76.0 \\
mAcc & 81.8 & \cellcolor[HTML]{def3e4}\bf 84.3 & 83.1 \\
\end{tabular}

        \vspace{-1mm}
        \subcaption{\textbf{Number of bottleneck layer.} The bottleneck layer often significantly affects model size but not always provide better performance.}
        \label{subtab:num_bottleneck}
        \vspace{1mm}
    \end{minipage}
    \hspace{0.03\textwidth}
    \begin{minipage}{0.26\textwidth}
    \centering
        \tablestyle{1pt}{1.08}        
        \begin{tabular}{x{15mm}|x{10mm}x{10mm}x{10mm}}
norm & BN & \cellcolor[HTML]{cce3de}LN \\\midrule
mIoU & 74.8 & \cellcolor[HTML]{def3e4}\bf 76.2 \\
mAcc &  82.7 & \cellcolor[HTML]{def3e4}\bf 84.3  \\
\end{tabular}

        \vspace{-1mm}
        \subcaption{\textbf{Normalization.} Batch norm is sensitive to batch size and degrade final performance compared to layer norm.
        }
        \label{subtab:ablation_norm}
        \vspace{1mm}
    \end{minipage}
        \hspace{0.03\textwidth}
    \begin{minipage}{0.28\textwidth}
    \centering
        \tablestyle{1pt}{1.08}
        \begin{tabular}{x{15mm}|x{10mm}x{10mm}x{10mm}}
LR Ramp \% & 5\% & \cellcolor[HTML]{cce3de}10\% \\\midrule
mIoU & 75.5 & \cellcolor[HTML]{def3e4}\bf 76.2 \\
mAcc & 82.7 & \cellcolor[HTML]{def3e4}\bf 84.3 \\
\end{tabular}

        \vspace{-1mm}
        \subcaption{\textbf{LR Warm-up Ratio}. Smaller warm-up ratios lead to training instability and degrade final performance.}
        \label{subtab:ablation_lr_warmup}
        \vspace{1mm}
    \end{minipage} \\
    \vspace{-5mm}
\end{table}

\paragraph{Ablations.}
We conducted comprehensive ablation studies to evaluate the contribution of each component to our model's performance (\Cref{tab:ablation}). 
As shown in \Cref{subtab:ablation_receptive_field,subtab:ablation_nrp}, both our receptive field calibration and the next-resolution prediction training are critical for cultivating resolution-invariant representations. To address data imbalance, we attempted to utilize focal loss\cite{lin2017focal} in \Cref{subtab:ablation_focal_loss} to dynamically adjust class weights. However, while this approach initially accelerated mIoU convergence, it ultimately led to lower final performance. Analysis reveals that the model tended to over-predict challenging classes, such as windows and pictures, thereby increasing false positives and degrading overall IoU.
Furthermore, \Cref{subtab:ablation_loss_weight} shows that assigning higher weights to higher-resolution prediction targets yields superior results, while \Cref{subtab:ablation_latent_space_alignment} confirms the effectiveness of our alignment loss (\Cref{eq:align}). Finally, we investigate hyperparameter sensitivity regarding bottleneck capacity, normalization techniques, and learning rate warm-up in \Cref{subtab:num_bottleneck,subtab:ablation_norm,subtab:ablation_lr_warmup}.

\section{Related Work}
\label{sec:related_work}
\vspace{-3mm}

\noindent{\bf Point Cloud Backbones and Local Neighborhoods.}
Point cloud backbones differ mainly in how they define the local neighborhood over which features are aggregated, and this choice determines how they respond to resolution shifts. PointNet~\cite{qi2017pointnet} processes raw coordinates with shared MLPs and aggregates them through global max-pooling, which makes it robust to missing input points. PointNet++~\cite{qi2017pointnet++} and PointNeXt~\cite{qian2022pointnext} group points hierarchically within a fixed metric radius, whereas DGCNN~\cite{wang2019dynamic} and Point Transformer v1/v2~\cite{zhao2021point, wu2022point} rely on $k$-nearest neighbors, so the physical extent of the neighborhood changes with density. KPConv~\cite{thomas2019kpconv} and sparse voxel convolutions~\cite{choy20194d} tie the kernel extent directly to the grid size. More recent backbones move toward efficient structured representations: OctFormer~\cite{wang2023octformer} applies attention within octree-based windows, PTv3~\cite{wu2024point} serializes points with space-filling curves into fixed-size patches, and LitePT~\cite{yuelitept2026} restricts attention to deep, low-resolution stages while using convolutions for early local feature extraction. In all of these designs, the neighborhood is defined by a fixed number of points, a fixed radius, or the grid size, so either its physical extent or the number of points it contains changes when the input resolution shifts (\Cref{tab:model_family}). Rather than proposing a new neighborhood definition, Invaria keeps a serialized transformer backbone and calibrates its receptive field to the input point cloud.

\noindent{\bf Scale- and Density-Invariant Features.}
Scale invariance has long been studied in 2D vision, where hand-crafted descriptors such as SIFT~\cite{lowe2004distinctive} and SURF~\cite{bay2006surf} detect keypoints in a scale space so that the same structure is described consistently across image scales. In deep point cloud learning, PointNet++~\cite{qi2017pointnet++} addresses non-uniform sampling density with multi-scale and multi-resolution grouping, and PointClustering~\cite{long2023pointclustering} learns representations that are invariant to transformations of the input during unsupervised pre-training. Sonata~\cite{wu2025sonata} further identifies a ``geometric shortcut'' in self-supervised point cloud learning, where models rely on low-level spatial cues rather than high-level semantics (see \Cref{sec:sup_related_work} for a broader overview of self-supervised pre-training). Compared to 2D vision, however, dedicated scale- and density-invariant representations for 3D point clouds remain scarce, and existing approaches mostly focus on invariant local descriptors or on invariance to transformations of the input. In contrast, Invaria targets scale and density robustness at the level of the backbone representation, through next-resolution prediction and receptive field calibration.

\noindent{\bf Next-Resolution Prediction}
The success of Large Language Models in next-token prediction~\cite{brown2020language, radford2018improving} has inspired a shift toward generative paradigms in visual and 3D domains. Recent frameworks have adopted next-resolution prediction. This approach, popularized by the Visual Autoregressive (VAR) modeling strategy~\cite{tian2024visual}, treats generation as a progressive refinement from coarse global structures to fine-grained geometric details. This transition to multi-scale tokenization is exemplified by Infinity~\cite{han2025infinity}, which introduces bitwise autoregressive modeling to scale high-resolution synthesis by predicting hierarchical bit-representations. In the 3D domain, PointNSP~\cite{meng2025pointnsp} and SAR3D~\cite{chen2025sar3d} adapt this philosophy by re-framing 3D generation as a sequence of nested-scale predictions. While these methods use coarse-to-fine prediction for generation, Invaria uses next-resolution prediction as a representation-learning objective: by predicting higher-resolution features from a lower-resolution input, the encoder is encouraged to learn semantic features that remain consistent across resolutions.

\section{Limitations}
\vspace{-3mm}
Invaria still struggles with simple planar objects such as doors and windows, which are easily confused with walls, and is affected by the class imbalance of existing 3D datasets. Moreover, on ScanNet and nuScenes, its accuracy drops sharply below roughly 5--6\% of the original points, where the geometry itself is largely lost. See \Cref{sec:limitation,sec:additional_exp} for details.

\section{Conclusion}
\label{sec:conclusion}
\vspace{-3mm}
We introduce Invaria, a 3D point cloud encoder that achieves scale and density invariance through next-resolution prediction and receptive field calibration. Our experiments demonstrate that while lower resolutions predictably degrade performance, higher-resolution inputs can paradoxically reduce accuracy as well. By extracting invariant features from sparser inputs, Invaria's asymmetric encoding strategy delivers robust performance across resolution shifts while significantly reducing computational overhead and model size. While the model currently excels at learning scale- and density-invariant features, exciting avenues for future work include better disambiguating simple planar geometries (such as differentiating doors or pictures from walls) and leveraging recent advances in generative modeling to address the severe class imbalances found in standard 3D datasets. 
\begin{ack}
The Cynergy4MIE project is supported by the Chips Joint Undertaking and its members, including the top-up funding by National Authorities under Grant Agreement No 101140226. 
This work has also been partially funded by the German Research Foundation (DFG) project ARROW (GA Nr 564809505) and by the EU project SUBLIMITY (GA Nr 101294363).
The authors declare no competing interests.
\end{ack}

{
    \small
    \bibliographystyle{unsrt}
    \bibliography{main}
}

%%%%%%%%%%%%%%%%%%%%%%%%%%%%%%%%%%%%%%%%%%%%%%%%%%%%%%%%%%%%
\newpage
\appendix
\section*{Appendix}
This appendix provides supplementary information to support the findings presented in the main paper. Specifically, we detail our experimental environment and data usage in \Cref{sec:env_data}. \Cref{sec:model_details} provides comprehensive architectural and hyperparameter configurations for Invaria. \Cref{sec:sup_related_work} reviews self-supervised pre-training for point clouds. Further discussion regarding the limitations of our method is included in \Cref{sec:limitation}, while \Cref{sec:additional_exp} presents additional empirical evaluations, including per-category performance breakdowns and robustness analyses. Finally, \Cref{sec:more_vis} offers supplementary qualitative visualizations to further illustrate the performance of our approach compared to existing state-of-the-art baselines. 

\section{Environment and Data Usage}
\label{sec:env_data}

\paragraph{Software Environment and Hardware Resources}
\begin{itemize}
    \item \textbf{OS:} Ubuntu 24.04.3 LTS (Noble Numbat)
    \item \textbf{Python version:} 3.10.19
    \item \textbf{PyTorch version:} 2.9.1+cu128 (built against CUDA 12.8)
    \item \textbf{Vision library:} torchvision 0.24.1
    \item \textbf{CUDA / cuDNN version:} 12.8 / 9.10.2
    \item \textbf{GPU:} NVIDIA A100 80GB $\times$ 2 for training; NVIDIA A100 80GB $\times$ 1 for evaluation.
\end{itemize}

\paragraph{Data Licenses} 
In our experiments, we utilize the publicly available datasets nuScenes \cite{caesar2020nuscenes}, Waymo~\cite{Sun_2020_CVPR}, DALES~\cite{varney2020dales}, and ScanNet \cite{dai2017scannet}. 
The nuScenes dataset is used under the Creative Commons Attribution-NonCommercial-ShareAlike 4.0 International (CC BY-NC-SA 4.0) license. 
The Waymo Open Dataset is used under the Waymo Dataset License Agreement for Non-Commercial Use. DALES is derived from aerial LiDAR data provided by the City of Surrey under the Open Government License -- City of Surrey.
The ScanNet dataset is provided under a custom Terms of Use from the Stanford, Princeton, and TUM organizers, which restricts usage to non-commercial research and educational purposes. 
We have strictly adhered to the licensing agreements and terms of use for all datasets.

\section{Model Details}
\label{sec:model_details}
The proposed Invaria architecture consists of three primary modules: an encoder, a decoder, and a predictor. The encoder and decoder adopt a U-Net architecture~\cite{ronneberger2015u}, building upon the implementation framework introduced by PTv3~\cite{wu2024point}. The predictor module integrate sparse convolutions and Multi-Layer Perceptrons, interspersed with normalization layers and skip connections to maintain feature consistency. 
We optimize the network parameters using the AdamW optimizer, employing a OneCycleLR learning rate scheduler to ensure stable convergence and efficient training.

\paragraph{Parameter Details.}
The encoder backbone is structured with five stages, using depths (2, 2, 2, 2, 1), channel widths (32, 64, 128, 256, 512), and attention heads (2, 4, 8, 16, 32). The decoder comprises four stages with depths (2, 2, 2, 2), channels (64, 64, 128, 256), and head configurations (4, 4, 8, 16), with a backbone output dimension of 64. Optimization is performed using AdamW with a base learning rate of 0.006 and a weight decay of 0.05. We employ the OneCycleLR scheduler with a peak learning rate of 0.006, an initial warm-up period of 10\% ($pct\_start=0.1$), and a cosine annealing strategy to ensure stable convergence.
For the scaling factor $\alpha$: We use $\alpha = 1.0$ for ScanNet and $\alpha = 0.8$ for nuScenes, Waymo, and DALES. Receptive field calibration is not sensitive to this choice: varying $\alpha$ within the range of 0.5--2 changes the mIoU by less than 0.5.

\paragraph{Invaria-s.}
Building upon the PTv3 backbone, LitePT~\cite{yuelitept2026} introduces a lightweight architecture optimized for computational efficiency without significantly compromising discriminative power. 
We integrate this design into our framework to develop Invaria-s, a reduced-capacity variant of our model. Invaria-s features a notable reduction in architectural footprint, with 16.3M parameters compared to the 25.5M in the baseline Invaria, alongside a $\sim$7\% reduction in FLOPs. 
While the results demonstrate promising enhancements in resolution-invariant evaluation, we observe a slight performance trade-off, as shown in \Cref{tab:resolution_change}, with a marginal drop of approximately 3--4\% in per-resolution accuracy. Nevertheless, this variant underscores the flexibility of our framework, providing a viable balance between computational efficiency and robustness for resource-constrained deployments.

\paragraph{Lower-Resolution Sampling.} 
During training, we simulate lower-resolution inputs by stochastically subsampling the original point cloud $\mathcal{P}$. Given an input with $N$ points, we define a set of target cardinalities $\mathcal{S} = \{n_0 \cdot h^i \mid i = 0, 1, \dots, k\}$, where $n_0$ is the base number of points and $h$ is the geometric scaling factor. For our experiments on ScanNet, we set $n_0 = 4096$ and $h = 4$, yielding $\mathcal{S} = \{4096, 4096 \cdot 4, 4096 \cdot 4^2\}$. We use the same $n_0 = 4096$ for all datasets. This training strategy forces the model to learn representations that are invariant to the underlying sampling density. At inference, we evaluate the model using subset cardinalities $n_i \in \mathcal{S}$, which are scaled relative to the input scene size $N$ while strictly maintaining the constraint $n_i < N$.

\begin{algorithm}[t]                         
  \caption{Receptive Field Calibration} 
  \label{alg:rfc}
  \begin{algorithmic}[1]
  \Require Point cloud $\mathcal{P} = \{x_j\}_{j=1}^{n_i}$;  anchor count $N_a$; scale factor $\alpha$; reduction $\rho$ 
  \Ensure  Calibrated grid size $g = \psi(\mathcal{P})$
  \State $\mathcal{P}_a \gets$ uniformly random subset of $\mathcal{P}$ with $|\mathcal{P}_a| = N_a$
  \For{each $a \in \mathcal{P}_a$}
      \State $d_a \gets \min_{x \in \mathcal{P} \setminus \{a\}} \|a - x\|$
      \Comment{nearest-neighbor distance}
  \EndFor
  \State $g \gets \alpha \cdot \rho\!\left(\{d_a\}_{a \in \mathcal{P}_a}\right)$
  \State \Return $g$
  \end{algorithmic}
\end{algorithm}

\paragraph{Receptive Field Calibration}
In \Cref{sec:method} we describe our Receptive Field Calibration; the detailed procedure is summarized in \Cref{alg:rfc}. The most intuitive choice for the reduction $\rho(\cdot)$ is the minimum, $\rho_{\min}(\{d_a\}) = \min_a d_a$, since by construction every pair of points must be separable by the resulting grid. In practice, however, $\rho_{\min}$ tends to severely underestimate the desired cell size. The reason is geometric: anchor distances are not measured between cell  centers but between arbitrary point locations, and since every interior cell has $26$ neighbors in $\mathbb{R}^3$, two points sitting near a shared face, edge, or corner can be arbitrarily close while still belonging to distinct cells. Therefore, our empirical findings suggest that $\rho_{\text{mean}}$ is a better and more stable choice.

\paragraph{Computational Complexity of Receptive Field Calibration.}
A full pairwise nearest-neighbor search over $N$ points costs $O(N^2)$. Instead, \Cref{alg:rfc} draws a fixed number of anchors $N_a$ (we use $N_a = 100$), independent of $N$, each requiring a single $O(N)$ scan, which yields a total cost of $O(N_a N) = O(N)$. The calibration is computed once per scene from the input coordinates alone and is therefore independent of the depth and width of the backbone. On an NVIDIA A100 GPU, it adds approximately 2\% to the end-to-end inference latency on ScanNet and nuScenes.
\section{Additional Related Work}
\label{sec:sup_related_work}

\paragraph{Self-supervised Pre-training and Representation Learning for Point Cloud}
Beyond architectural refinements, 3D point cloud understanding has shifted toward Self-Supervised Learning (SSL)~\cite{jaiswal2020survey, liu2021self,zeng2024self, abdelsamad2025multi} to extract semantically rich representations from unlabeled data. Sonata~\cite{wu2025sonata} mitigates ``geometric shortcuts'', where models prioritized low-level spatial patterns over high-level concepts, via a multi-scale self-distillation framework that obscures spatial details to force the learning of robust semantic features. To further enrich these representations, Concerto~\cite{zhang2025concerto} leverages multi-sensory synergy through 2D-3D cross-modal joint embedding. Building on these foundational trends, Utonia~\cite{zhang2026utonia} further employs a unified point transformer trained across heterogeneous domains, ranging from indoor to outdoor.

\section{Limitation and Discussion}
\label{sec:limitation}
\paragraph{Ambiguity in Planar Objects.}
While mIoU provides a high-level summary of model performance, it often masks significant disparities between semantic categories. A granular per-class analysis reveals that objects with simple, planar geometries, such as door and window, pose a distinct challenge~\cite{chang2024mikasa, chang20253d}, as the model frequently conflates these structures with walls and floors due to their shared, near-2D geometric profile. Notably, the floor category exhibits notable resilience. This robustness is mainly attributable to the vertical Z-coordinate, which provides a strong spatial prior that facilitates disambiguation, a structural cue absent in other simple planar objects.

\paragraph{Class Imbalance.}
As illustrated in \Cref{tab:scannet_dis}, existing 3D benchmarks, including ScanNet~\cite{dai2017scannet}, nuScenes~\cite{caesar2020nuscenes}, and Waymo~\cite{Sun_2020_CVPR}, suffer from inherent class imbalance. This imbalance also affects robustness to resolution shifts: \textit{Wall} and \textit{Floor} together account for over 55\% of the labeled points in ScanNet, whereas fragile classes such as \textit{Door} (4.5\%) and \textit{Shower Curtain} (0.4\%) are far less frequent, which limits the geometric cues the model can learn for them. While prior literature often addresses this bias by reweighting the loss function based on class frequency (e.g., via Focal Loss~\cite{lin2017focal}), our ablation study (\Cref{subtab:ablation_focal_loss}) demonstrates that this approach yields no performance gains for our framework. Observations of the training dynamics reveal a competitive interplay between classes, where improvements in one category appear to necessitate trade-offs in others, suggesting conflicting optimization objectives~\cite{chang2025seeing, chang2026probing}. Leveraging recent advancements in generative modeling, future research could potentially mitigate these persistent distribution biases by augmenting existing datasets with high-fidelity, synthetic samples of underrepresented classes.

\section{Additional Results}
\label{sec:additional_exp}
\paragraph{Sensitivity to Density Variations.}
As discussed in \Cref{sec:cause}, point cloud encoders often exploit density as a "shortcut" when learning semantic features. To quantify the robustness of existing methods to such variations, we evaluate their performance under varying sampling densities. Theoretically, increasing the grid size $\tau$ by a factor of $k$ reduces the number of points $N$ proportional to $1/k^3$; consequently, a 2- to 3-fold increase in $\tau$ should theoretically result in an 8- to 27-fold reduction in density. Here we empirically evaluate model robustness at 10x and 20x reduced sampling rates. 

The detailed results are provided in \Cref{tab:density_change}, showing our proposed Invaria can stay more robust when facing sparser input compared to other existing works.

\paragraph{Detailed Per-Category Analysis.}
We provide a comprehensive performance comparison across categories in ~\Cref{tab:resolution_details} (resolution shifts on ScanNet), ~\Cref{tab:nuscenes_resolution_detail} (resolution shifts on nuScenes), and ~\Cref{tab:scale_details} (scale changes on ScanNet). As noted in \Cref{sec:limitation}, while planar objects typically exhibit lower baseline performance, complex geometries demonstrate a more significant performance degradation as resolution or scale varies across existing state-of-the-art methods.

\begin{table}[t]
\centering
\footnotesize
\caption{Robustness to resolution shifts on outdoor LiDAR datasets, reported in mIoU (\%). \textit{Ret.}: mIoU at $3\tau^*$ relative to the mIoU at the default resolution $\tau^*$.}
\label{tab:outdoor_overall}
\setlength{\tabcolsep}{3.0pt}
\begin{tabular}{l cccc cccc cccc}
\toprule
 \multirow{2}{*}{\textbf{Model}} & \multicolumn{4}{c}{\textbf{nuScenes} (32-beam)} & \multicolumn{4}{c}{\textbf{DALES} (aerial)} & \multicolumn{4}{c}{\textbf{Waymo} (64-beam)} \\
\cmidrule(lr){2-5} \cmidrule(lr){6-9} \cmidrule(lr){10-13}
 & \cellcolor[HTML]{D3D3D3}$\tau^*$ & $2\tau^*$ & $3\tau^*$ & Ret. & \cellcolor[HTML]{D3D3D3}$\tau^*$ & $2\tau^*$ & $3\tau^*$ & Ret. & \cellcolor[HTML]{D3D3D3}$\tau^*$ & $2\tau^*$ & $3\tau^*$ & Ret. \\
\midrule
Sp.Mink.\cite{choy20194d} & 76.7 & 35.4 & 14.5 & 18.9\% & 80.2 & 53.2 & 32.9 & 41.0\% & 69.8 & 44.5 & 23.7 & 34.0\% \\
PTv3\cite{wu2024point} & \bf 80.4 & 51.9 & 26.4 & 32.9\% & \bf 81.7 & 59.6 & 37.2 & 45.5\% & \bf 71.0 & 54.0 & 31.7 & 44.6\% \\
Invaria & 76.2 & \bf 75.6 & \bf 71.9 & \bf 94.4\% & 80.2 & \bf 80.1 & \bf 76.3 & \bf 95.2\% & 70.6 & \bf 69.7 & \bf 66.0 & \bf 93.5\% \\
\bottomrule
\end{tabular}
\end{table}

\begin{table}[t]
\centering
\caption{Per-class IoU (\%) on outdoor LiDAR datasets. Each cell reports the IoU at the default resolution $\tau^*$ $\rightarrow$ the IoU at $3\tau^*$. Classes are ordered from spatially extended (left) to compact (right).}
\label{tab:outdoor_per_class}
\resizebox{\textwidth}{!}{
\setlength{\tabcolsep}{2.5pt}
\begin{tabular}{l cccccccccc}
\toprule
\textbf{nuScenes} & driveable surface & sidewalk & vegetation & manmade & truck & car & bicycle & pedestrian & traffic cone & mIoU \\
\midrule
Sp.Mink.\cite{choy20194d} & 96.7$\to$55.9 & 75.6$\to$19.1 & 86.1$\to$51.1 & 88.8$\to$51.8 & 86.5$\to$0.8 & 86.9$\to$0.2 & 37.1$\to$0.0 & 80.1$\to$1.1 & 64.0$\to$11.9 & 76.7$\to$14.5 \\
PTv3\cite{wu2024point} & 97.1$\to$84.7 & 77.1$\to$38.6 & 89.5$\to$73.8 & 90.9$\to$66.0 & 84.1$\to$16.6 & 92.3$\to$13.8 & 50.7$\to$0.0 & 82.9$\to$2.0 & 71.0$\to$18.5 & 80.4$\to$26.4 \\
Invaria & 96.7$\to$96.1 & 75.6$\to$71.6 & 87.7$\to$86.4 & 89.3$\to$88.0 & 83.8$\to$79.2 & 91.8$\to$89.0 & 26.9$\to$20.2 & 77.6$\to$68.3 & 65.2$\to$56.4 & 76.2$\to$71.9 \\
\midrule
\textbf{DALES} & ground & buildings & vegetation & power lines & fences & poles & cars & trucks & & mIoU \\
\midrule
Sp.Mink.\cite{choy20194d} & 97.1$\to$87.6 & 96.7$\to$57.0 & 93.9$\to$65.7 & 92.2$\to$17.7 & 63.2$\to$14.0 & 69.4$\to$20.7 & 85.0$\to$0.6 & 43.8$\to$0.01 & & 80.2$\to$32.9 \\
PTv3\cite{wu2024point} & 97.1$\to$92.4 & 97.1$\to$67.1 & 93.7$\to$69.3 & 94.7$\to$26.0 & 66.2$\to$26.4 & 74.8$\to$16.2 & 85.6$\to$0.02 & 44.4$\to$0.00 & & 81.7$\to$37.2 \\
Invaria & 96.5$\to$96.0 & 96.8$\to$96.1 & 92.9$\to$91.8 & 89.1$\to$85.5 & 63.9$\to$54.5 & 73.5$\to$69.5 & 84.7$\to$77.8 & 44.3$\to$39.6 & & 80.2$\to$76.3 \\
\midrule
\textbf{Waymo} & road & sidewalk & vegetation & building & truck & car & pole & pedestrian & construction cone & mIoU \\
\midrule
Sp.Mink.\cite{choy20194d} & 93.1$\to$78.6 & 77.7$\to$32.1 & 91.1$\to$68.3 & 95.3$\to$66.9 & 60.9$\to$7.5 & 94.6$\to$19.2 & 74.8$\to$34.1 & 89.4$\to$2.2 & 70.8$\to$28.5 & 69.8$\to$23.7 \\
PTv3\cite{wu2024point} & 93.4$\to$85.3 & 77.7$\to$43.3 & 91.6$\to$80.8 & 95.8$\to$73.5 & 63.8$\to$14.5 & 94.5$\to$29.7 & 77.3$\to$40.9 & 90.1$\to$3.4 & 72.6$\to$37.7 & 71.0$\to$31.7 \\
Invaria & 93.3$\to$92.8 & 77.6$\to$76.0 & 91.4$\to$90.6 & 95.7$\to$94.8 & 64.2$\to$46.1 & 94.8$\to$95.1 & 77.1$\to$73.8 & 89.8$\to$87.8 & 72.2$\to$73.5 & 70.6$\to$66.0 \\
\bottomrule
\end{tabular}
}
\end{table}

\paragraph{Additional Outdoor Datasets.}
To verify that our findings are not specific to ScanNet, whose point clouds are derived from reconstructed meshes, we additionally evaluate on DALES~\cite{varney2020dales}, an aerial LiDAR dataset, alongside the outdoor driving datasets nuScenes~\cite{caesar2020nuscenes} and Waymo~\cite{Sun_2020_CVPR}. Unlike ScanNet, which is captured indoors with a handheld RGB-D sensor, DALES covers large-scale terrain and infrastructure scanned from an airborne, top-down viewpoint, which also distinguishes it from the ground-level driving scenes of nuScenes and Waymo. All three outdoor datasets consist of raw LiDAR measurements and therefore naturally contain real sensor artifacts such as occlusion, range-dependent noise, density fall-off, and missing returns. As shown in \Cref{tab:outdoor_overall}, the baselines lose most of their accuracy under a $3\times$ resolution shift on all three datasets, whereas Invaria consistently retains over 93\% of its performance. The failure mode we identify is therefore not limited to ScanNet or to a single sensor configuration.

\definecolor{lightgray}{gray}{0.9}
\begin{table}[t]
\centering
\footnotesize % Give it a starting point so resizebox doesn't work as hard
\begin{minipage}{0.48\textwidth}
    \centering
    \caption{The density change, evaluating on sparser input point cloud.}
    \label{tab:density_change}
    \resizebox{\textwidth}{!}{
        \setlength{\tabcolsep}{1.5pt} % CRITICAL: Tighten spacing
        
\begin{tabular}{l cc p{0.1cm} cc p{0.1cm} cc}
\toprule
 \multirow{2}{*}{\textbf{Model}} & \multicolumn{2}{c}{\textbf{mIoU (\%)}} & & \multicolumn{2}{c}{\textbf{mAcc (\%)}} & & \multicolumn{2}{c}{\textbf{allAcc (\%)}} \\
\cmidrule{2-3} \cmidrule{5-6} \cmidrule{8-9}
 & $5\%$ & $10\%$  & & $5\%$ & $10\%$ & & $5\%$ & $10\%$ \\
\midrule
Sonata      & 36.6 & 60.4 && 45.0 & 69.8 && 65.4 & 82.1 \\
Concrt-L    & 39.9 & \cellcolor[HTML]{def3e4}\bf65.4 && 49.7 & \cellcolor[HTML]{def3e4}\bf73.8 && 67.1 & \cellcolor[HTML]{eef9f0}84.5 \\
Utonia      & 41.8 & 63.9 && 50.8 & 72.5 && 68.2 & 83.5 \\
\midrule
Sp.Mink.    & 13.5 & 30.8 && 20.3 & 37.3 && 60.8 & 74.0 \\
PTv3        & 24.8 & 61.7 && 30.0 & 69.4 && \cellcolor[HTML]{eef9f0}68.9 & \cellcolor[HTML]{def3e4}\bf85.5 \\
Invaria-s   & \cellcolor[HTML]{eef9f0}43.4 & 61.9 && \cellcolor[HTML]{eef9f0}53.0 & 69.9 && 68.3 & 82.2 \\
Invaria     & \cellcolor[HTML]{def3e4}\bf48.2 & \cellcolor[HTML]{eef9f0}64.0 && \cellcolor[HTML]{def3e4}\bf56.4 & \cellcolor[HTML]{eef9f0}72.7 && \cellcolor[HTML]{def3e4}\bf71.2 & 83.1 \\

\bottomrule
\end{tabular}

    }
\end{minipage}
\hfill
\begin{minipage}{0.5\textwidth}
    \centering
    \caption{The distribution of each categories in ScanNet~\cite{dai2017scannet}, Showing imbalance in different classes. }
    \label{tab:scannet_dis}
    \resizebox{\textwidth}{!}{
        \setlength{\tabcolsep}{1.5pt}
        \begin{tabular}{l r r r r r}
        \toprule
        Class & Count & \% Valid & Cum \% & Scenes & \% Scenes \\
        \midrule
        Wall & 10,936,492 & 30.48\% & 30.48\% & 305 & 97.8\% \\
        Floor & 8,983,296 & 25.04\% & 55.52\% & 312 & 100.0\% \\
        Chair & 3,194,103 & 8.90\% & 64.42\% & 220 & 70.5\% \\
        Other Furniture & 1,637,177 & 4.56\% & 68.98\% & 240 & 76.9\% \\
        Door & 1,600,191 & 4.46\% & 73.44\% & 219 & 70.2\% \\
        Table & 1,478,887 & 4.12\% & 77.56\% & 178 & 57.1\% \\
        Cabinet & 1,399,895 & 3.90\% & 81.46\% & 139 & 44.6\% \\
        Window & 1,385,131 & 3.86\% & 85.32\% & 166 & 53.2\% \\
        Bookshelf & 1,125,319 & 3.14\% & 88.46\% & 46 & 14.7\% \\
        Bed & 915,295 & 2.55\% & 91.01\% & 61 & 19.6\% \\
        Sofa & 789,769 & 2.20\% & 93.21\% & 69 & 22.1\% \\
        Curtain & 657,363 & 1.83\% & 95.04\% & 48 & 15.4\% \\
        Desk & 655,163 & 1.83\% & 96.87\% & 84 & 26.9\% \\
        Refrigerator & 223,258 & 0.62\% & 97.49\% & 46 & 14.7\% \\
        Counter & 221,560 & 0.62\% & 98.11\% & 47 & 15.1\% \\
        Picture & 207,442 & 0.58\% & 98.69\% & 90 & 28.8\% \\
        Shower Curtain & 158,208 & 0.44\% & 99.13\% & 27 & 8.7\% \\
        Toilet & 113,661 & 0.32\% & 99.45\% & 51 & 16.3\% \\
        Sink & 102,489 & 0.29\% & 99.73\% & 80 & 25.6\% \\
        Bathtub & 96,355 & 0.27\% & 100.00\% & 31 & 9.9\% \\
        \bottomrule
    \end{tabular}
    }
\end{minipage}
\end{table}
\paragraph{Per-Class Analysis on Outdoor Datasets.}
\Cref{tab:outdoor_per_class} reports the per-class IoU on the outdoor datasets. As on ScanNet, the sensitivity of a class follows its geometry. For PTv3, large planar surfaces are the most robust (87--95\% retention on driveable surface, ground, and road), volumetric classes such as vegetation, buildings, and manmade structures lie in the middle (69--88\%), and thin structures are fragile (e.g., 22\%, 27\%, and 40\% on poles, power lines, and fences in DALES). However, the degradation is far more severe outdoors, where compact classes collapse almost entirely: on DALES, PTv3 drops from 85.6 to 0.02 IoU on cars and from 44.4 to 0.00 on trucks, and on nuScenes from 50.7 to 0.0 on bicycles and from 82.9 to 2.0 on pedestrians. In contrast, Invaria remains robust on these classes, e.g., 84.7 $\to$ 77.8 on DALES cars and 89.8 $\to$ 87.8 on Waymo pedestrians.

\paragraph{Stress Test under Stronger Resolution Shifts.}
To reveal failure cases, we further reduce the input resolution beyond the range covered in \Cref{tab:resolution_change}. The degradation is not uniform: it remains gradual from 100\% down to about 11\% of the original point count, but on ScanNet and nuScenes, accuracy drops sharply once the input falls below roughly 5--6\% of the original points. We consider this a different failure mode from the one targeted in this work. The degradation addressed by Invaria is recoverable because the geometry is still present in the input, whereas below roughly 5--6\% of the original points, the geometry itself is largely lost.

\begin{sidewaystable}[p] % This rotates the table 90 degrees on the page
\centering
\caption{Detailed analysis of each category in ScanNet as the resolution changes. ($\tau^*=2 $cm)}
\label{tab:resolution_details}
\tiny % Extremely small font to fit all columns
\setlength{\tabcolsep}{1.2pt} % Tighten column spacing
\renewcommand{\arraystretch}{1.2} % Vertical breathing room
\begin{tabular}{l  l ccc *{20}{cc}}
\toprule
\textbf{Grid Size} & \textbf{Method} & \textbf{mIoU} & \textbf{mAcc} & \textbf{allAcc} & 
\multicolumn{2}{c}{Wall} & \multicolumn{2}{c}{Floor} & \multicolumn{2}{c}{Cab.} & 
\multicolumn{2}{c}{Bed} & \multicolumn{2}{c}{Chair} & \multicolumn{2}{c}{Sofa} & 
\multicolumn{2}{c}{Tabl} & \multicolumn{2}{c}{Door} & \multicolumn{2}{c}{Wind} & 
\multicolumn{2}{c}{Bksh} & \multicolumn{2}{c}{Pict} & \multicolumn{2}{c}{Cntr} & 
\multicolumn{2}{c}{Desk} & \multicolumn{2}{c}{Curt} & \multicolumn{2}{c}{Frid} & 
\multicolumn{2}{c}{Shw.} & \multicolumn{2}{c}{Toil} & \multicolumn{2}{c}{Sink} & 
\multicolumn{2}{c}{Bath} & \multicolumn{2}{c}{Oth.} \\

\cmidrule(lr){6-7} \cmidrule(lr){8-9} \cmidrule(lr){10-11} \cmidrule(lr){12-13} \cmidrule(lr){14-15} \cmidrule(lr){16-17} \cmidrule(lr){18-19} \cmidrule(lr){20-21} \cmidrule(lr){22-23} \cmidrule(lr){24-25} \cmidrule(lr){26-27} \cmidrule(lr){28-29} \cmidrule(lr){30-31} \cmidrule(lr){32-33} \cmidrule(lr){34-35} \cmidrule(lr){36-37} \cmidrule(lr){38-39} \cmidrule(lr){40-41} \cmidrule(lr){42-43} \cmidrule(lr){44-45}

& & (\%) & (\%) & (\%) & IoU & Acc & IoU & Acc & IoU & Acc & IoU & Acc & IoU & Acc & IoU & Acc & IoU & Acc & IoU & Acc & IoU & Acc & IoU & Acc & IoU & Acc & IoU & Acc & IoU & Acc & IoU & Acc & IoU & Acc & IoU & Acc & IoU & Acc & IoU & Acc & IoU & Acc & IoU & Acc \\
\midrule

% --- BLOCK 1: x0.5 ---
\multirow{7}{*}{$0.5\tau^* $} 
& Sonata   & 59.9 & 70.7 & 84.9 & 82.1 & 94.9 & 94.1 & 97.9 & 51.7 & 67.5 & 62.4 & 81.6 & 76.9 & 80.3 & 56.8 & 82.3 & 57.1 & 74.3 & 52.7 & 63.5 & 58.5 & 69.5 & 65.0 & 75.4 & 37.0 & 44.8 & 48.3 & 64.5 & 35.3 & 42.6 & 78.0 & 88.6 & 37.5 & 46.3 & 69.3 & 74.4 & 66.9 & 68.6 & 50.6 & 57.4 & 76.1 & 85.1 & 42.3 & 55.3 \\
& Concrt-L & 67.8 & 77.5 & 88.4 & 85.0 & 95.0 & 94.8 & 97.6 & 63.1 & 79.2 & 72.0 & 89.2 & 80.6 & 84.6 & 62.9 & 85.6 & 65.2 & 78.5 & 60.6 & 68.2 & 71.7 & 82.9 & 77.0 & 83.7 & 35.0 & 41.1 & 57.8 & 71.7 & 50.5 & 58.5 & 75.7 & 90.5 & 63.3 & 70.5 & 44.5 & 47.0 & 86.3 & 88.4 & 65.7 & 73.2 & 84.9 & 90.0 & 58.8 & 73.7 \\
& Utonia   & 73.0 & 81.9 & 95.4 & 84.8 & 95.4 & 95.2 & 97.6 & 67.4 & 77.3 & 78.2 & 88.5 & 83.4 & 86.2 & 65.0 & 89.9 & 72.4 & 82.1 & 69.5 & 77.3 & 70.2 & 78.5 & 79.5 & 87.0 & 37.5 & 46.5 & 62.0 & 79.0 & 62.3 & 74.6 & 79.9 & 88.7 & 72.3 & 78.5 & 71.6 & 76.3 & 91.9 & 93.7 & 70.0 & 77.3 & 84.8 & 89.4 & 61.1 & 74.0 \\
& Sp.Mink. & 36.6 & 47.2 & 73.8 & 73.0 & 95.7 & 92.5 & 98.4 & 31.8 & 52.2 & 41.9 & 81.3 & 37.9 & 39.1 & 24.1 & 44.0 & 41.0 & 60.0 & 25.9 & 30.1 & 39.9 & 49.9 & 42.4 & 47.6 & 19.8 & 24.4 & 22.4 & 28.1 & 16.6 & 20.7 & 52.2 & 70.8 & 9.3 & 10.8 & 35.3 & 34.1 & 28.9 & 30.4 & 26.0 & 26.9 & 48.6 & 63.5 & 22.0 & 35.0 \\
& PTv3 & 50.4 & 59.3 & 80.6 & 76.4 & 97.0 & 90.0 & 98.9 & 45.7 & 65.8 & 59.4 & 80.2 & 70.5 & 73.2 & 39.9 & 56.3 & 48.6 & 71.4 & 40.1 & 45.2 & 48.2 & 56.6 & 59.2 & 62.1 & 22.9 & 24.7 & 25.8 & 28.1 & 25.5 & 27.1 & 61.3 & 75.4 & 33.2 & 35.4 & 46.7 & 49.1 & 66.2 & 67.3 & 49.0 & 54.2 & 69.8 & 81.6 & 30.7 & 35.7 \\
& Invaria & 75.5 & 83.8 & 91.0 & 86.4 & 94.9 & 95.6 & 98.4 & 70.0 & 81.6 & 83.4 & 88.8 & 91.6 & 95.5 & 84.0 & 93.8 & 74.2 & 81.8 & 67.5 & 76.9 & 68.0 & 81.4 & 82.7 & 92.0 & 37.8 & 43.2 & 65.1 & 75.7 & 65.6 & 89.7 & 78.0 & 86.1 & 74.7 & 82.0 & 71.4 & 77.3 & 95.4 & 97.7 & 69.0 & 77.7 & 89.1 & 94.0 & 61.2 & 67.1 \\
\midrule

% --- BLOCK 2: x1 ---
\multirow{7}{*}{$\tau^* $} 
& Sonata   & 72.6 & 83.4 & 89.8 & 85.7 & 93.0 & 95.1 & 97.8 & 63.2 & 75.8 & 78.5 & 84.8 & 88.0 & 93.2 & 79.3 & 91.0 & 71.5 & 81.7 & 67.7 & 81.1 & 68.2 & 84.4 & 77.2 & 92.4 & 37.6 & 49.3 & 63.7 & 80.3 & 62.7 & 86.0 & 81.5 & 87.8 & 57.0 & 74.8 & 73.2 & 80.2 & 91.2 & 96.9 & 67.7 & 80.8 & 85.8 & 94.1 & 55.0 & 62.2 \\
& Concrt-L & 78.6 & 87.4 & 92.3 & 87.4 & 93.0 & 95.5 & 97.7 & 72.3 & 82.7 & 82.3 & 89.9 & 90.7 & 95.5 & 81.8 & 90.7 & 78.5 & 86.8 & 77.5 & 91.7 & 79.1 & 89.4 & 86.7 & 94.2 & 37.6 & 53.7 & 68.8 & 85.6 & 68.6 & 80.5 & 86.0 & 91.1 & 79.8 & 89.9 & 74.9 & 80.3 & 94.0 & 96.6 & 72.0 & 82.8 & 86.9 & 94.3 & 71.9 & 81.8 \\
& Utonia   & 76.6 & 85.1 & 94.6 & 85.8 & 94.6 & 95.4 & 97.9 & 72.0 & 82.0 & 81.0 & 87.1 & 88.8 & 91.8 & 73.8 & 91.0 & 77.0 & 83.8 & 75.1 & 85.4 & 73.1 & 82.8 & 82.7 & 92.4 & 39.0 & 50.5 & 65.2 & 86.3 & 67.9 & 79.0 & 83.3 & 88.0 & 80.0 & 86.6 & 75.1 & 80.7 & 94.0 & 96.0 & 70.2 & 79.5 & 85.3 & 90.4 & 66.6 & 75.8 \\
& Sp.Mink. & 75.6 & 83.5 & 91.5 & 87.2 & 95.7 & 95.9 & 98.2 & 69.9 & 80.6 & 84.6 & 89.9 & 92.4 & 95.7 & 84.7 & 93.3 & 76.2 & 84.9 & 70.0 & 80.2 & 68.4 & 81.5 & 81.7 & 92.2 & 37.8 & 44.2 & 66.0 & 80.9 & 68.1 & 86.1 & 77.9 & 86.4 & 65.9 & 72.8 & 71.8 & 76.4 & 91.0 & 96.6 & 67.5 & 72.1 & 88.8 & 91.4 & 65.6 & 70.1 \\
& PTv3 & 77.5 & 85.0 & 92.0 & 87.2 & 95.7 & 95.7 & 98.3 & 71.2 & 83.1 & 83.4 & 87.8 & 92.9 & 96.4 & 86.3 & 93.6 & 79.0 & 86.6 & 74.9 & 84.4 & 74.0 & 84.1 & 83.0 & 91.6 & 39.8 & 48.5 & 71.7 & 81.7 & 73.9 & 89.0 & 79.1 & 87.5 & 71.3 & 75.7 & 72.2 & 77.3 & 93.4 & 97.9 & 71.0 & 81.5 & 88.2 & 91.4 & 63.0 & 66.7 \\
& Invaria & 76.2 & 84.3 & 91.2 & 86.6 & 94.9 & 95.7 & 98.4 & 71.2 & 82.0 & 83.6 & 89.0 & 91.9 & 95.7 & 84.6 & 93.9 & 74.2 & 82.4 & 69.1 & 78.7 & 69.9 & 82.6 & 83.1 & 92.6 & 38.4 & 44.0 & 65.5 & 76.6 & 65.4 & 89.3 & 79.8 & 86.7 & 75.5 & 83.2 & 74.2 & 80.0 & 95.7 & 97.8 & 69.8 & 78.3 & 89.6 & 94.2 & 60.1 & 66.1 \\
\midrule

% --- BLOCK 3: x2 ---
\multirow{7}{*}{$2\tau^* $} 
& Sonata   & 62.3 & 73.8 & 85.7 & 81.0 & 90.6 & 93.3 & 98.5 & 53.9 & 67.7 & 72.2 & 81.7 & 79.3 & 88.3 & 69.6 & 75.4 & 64.0 & 72.3 & 55.8 & 69.2 & 61.6 & 80.0 & 70.5 & 90.9 & 25.7 & 32.5 & 52.9 & 58.6 & 51.9 & 82.5 & 75.5 & 84.6 & 42.5 & 56.5 & 69.6 & 79.7 & 65.0 & 70.0 & 50.8 & 67.9 & 71.3 & 78.3 & 39.7 & 49.9 \\
& Concrt-L & 72.5 & 82.3 & 89.9 & 84.0 & 90.4 & 94.0 & 98.1 & 67.5 & 79.5 & 81.7 & 88.3 & 86.0 & 91.9 & 83.6 & 87.4 & 73.0 & 79.3 & 67.7 & 85.4 & 73.7 & 89.0 & 81.6 & 95.3 & 33.6 & 50.4 & 63.9 & 75.3 & 63.5 & 80.5 & 83.4 & 91.2 & 68.2 & 79.3 & 73.0 & 80.0 & 66.9 & 68.0 & 62.5 & 75.5 & 80.5 & 86.2 & 62.5 & 74.9 \\
& Utonia   & 71.7 & 80.5 & 89.2 & 82.8 & 93.6 & 93.9 & 98.2 & 66.5 & 79.9 & 79.6 & 83.9 & 86.1 & 90.8 & 79.7 & 88.9 & 71.7 & 76.3 & 68.7 & 80.1 & 68.0 & 81.5 & 76.4 & 83.4 & 28.8 & 35.7 & 63.3 & 79.7 & 65.1 & 78.1 & 77.1 & 84.4 & 64.7 & 74.2 & 72.3 & 80.5 & 87.7 & 91.4 & 63.7 & 74.7 & 80.4 & 86.3 & 57.4 & 68.1 \\
& Sp.Mink. & 47.8 & 56.0 & 82.4 & 78.0 & 94.9 & 93.9 & 98.3 & 45.5 & 61.4 & 57.5 & 59.7 & 75.8 & 84.9 & 53.5 & 60.1 & 60.1 & 76.3 & 39.5 & 49.7 & 56.9 & 73.6 & 69.2 & 79.8 & 16.4 & 17.5 & 44.5 & 52.7 & 44.2 & 57.8 & 66.3 & 80.5 & 25.2 & 25.8 & 37.2 & 38.7 & 8.8 & 9.2 & 28.7 & 31.3 & 25.2 & 26.8 & 29.2 & 41.5 \\
& PTv3 & 60.4 & 69.0 & 86.0 & 80.6 & 95.5 & 92.8 & 98.7 & 59.4 & 71.8 & 67.5 & 71.0 & 80.5 & 88.1 & 59.1 & 69.5 & 67.9 & 79.0 & 52.3 & 60.0 & 62.4 & 77.4 & 77.6 & 86.4 & 28.8 & 33.4 & 59.1 & 66.2 & 45.4 & 57.6 & 72.8 & 82.8 & 38.4 & 40.2 & 50.9 & 54.3 & 54.6 & 54.9 & 55.3 & 66.1 & 60.8 & 77.1 & 42.0 & 50.6 \\
& Invaria & 75.7 & 84.0 & 90.9 & 85.9 & 94.3 & 95.3 & 98.5 & 68.9 & 81.6 & 83.4 & 86.8 & 91.5 & 95.2 & 84.0 & 95.0 & 74.3 & 82.3 & 65.4 & 75.9 & 68.3 & 83.0 & 83.2 & 93.0 & 34.7 & 40.3 & 66.4 & 78.6 & 65.5 & 88.5 & 79.5 & 86.2 & 74.8 & 81.1 & 76.4 & 83.0 & 95.3 & 97.2 & 69.8 & 78.7 & 89.0 & 94.1 & 61.8 & 66.8 \\
\midrule
% --- BLOCK 4: x3 ---
\multirow{7}{*}{$3\tau^* $} 
& Sonata   & 34.2 & 43.6 & 72.3 & 71.5 & 92.0 & 88.6 & 98.2 & 28.2 & 40.7 & 43.8 & 52.4 & 42.2 & 49.3 & 20.1 & 21.3 & 40.8 & 54.6 & 23.1 & 28.3 & 46.6 & 62.2 & 45.6 & 75.3 & 10.9 & 12.0 & 7.9 & 8.6 & 24.0 & 31.9 & 60.0 & 69.7 & 15.9 & 18.6 & 57.0 & 71.9 & 8.7 & 10.6 & 14.9 & 22.5 & 23.0 & 29.6 & 10.9 & 23.1 \\
& Concrt-L & 46.0 & 57.3 & 79.1 & 76.1 & 87.7 & 89.9 & 98.0 & 39.6 & 63.9 & 67.2 & 74.1 & 59.4 & 65.3 & 68.4 & 73.5 & 52.7 & 64.5 & 46.2 & 62.5 & 58.7 & 83.4 & 46.8 & 69.8 & 23.8 & 42.8 & 26.6 & 33.7 & 35.8 & 51.1 & 70.1 & 83.7 & 26.3 & 28.7 & 57.6 & 62.5 & 3.5 & 3.6 & 20.1 & 30.9 & 28.3 & 30.3 & 22.8 & 35.6 \\
& Utonia   & 47.2 & 56.3 & 78.7 & 74.5 & 93.5 & 90.0 & 98.1 & 44.2 & 58.1 & 63.9 & 65.1 & 62.0 & 66.6 & 66.0 & 76.4 & 53.4 & 64.3 & 41.7 & 49.1 & 53.5 & 71.7 & 19.2 & 20.2 & 14.8 & 17.4 & 37.2 & 49.1 & 35.0 & 40.1 & 63.9 & 76.0 & 31.3 & 33.5 & 54.9 & 61.0 & 26.5 & 27.3 & 38.8 & 46.5 & 47.1 & 57.6 & 26.8 & 54.4 \\
& Sp.Mink. & 19.4 & 25.7 & 66.1 & 64.8 & 95.4 & 85.0 & 95.5 & 20.9 & 31.7 & 7.4 & 7.5 & 35.5 & 40.1 & 9.2 & 9.3 & 37.5 & 56.0 & 8.8 & 10.7 & 33.0 & 52.0 & 23.8 & 26.6 & 0.6 & 0.6 & 0.0 & 0.0 & 5.6 & 7.2 & 45.9 & 56.3 & 0.0 & 0.0 & 0.2 & 0.2 & 0.0 & 0.0 & 0.4 & 0.4 & 0.0 & 0.0 & 9.8 & 23.7 \\
& PTv3 & 20.5 & 26.0 & 65.9 & 60.2 & 96.4 & 74.4 & 99.0 & 13.0 & 15.0 & 11.9 & 12.0 & 29.6 & 34.0 & 4.2 & 4.2 & 35.7 & 43.5 & 19.6 & 22.1 & 41.1 & 62.3 & 26.8 & 27.7 & 5.9 & 7.9 & 0.1 & 0.1 & 3.1 & 3.2 & 47.7 & 51.3 & 0.1 & 0.1 & 12.4 & 12.8 & 0.0 & 0.0 & 4.6 & 4.8 & 10.6 & 11.8 & 8.4 & 11.1 \\
& Invaria & 73.8 & 82.7 & 90.0 & 84.5 & 94.0 & 95.1 & 98.5 & 67.9 & 80.9 & 82.0 & 86.8 & 90.1 & 94.2 & 82.8 & 95.3 & 73.7 & 81.3 & 59.7 & 70.3 & 65.8 & 80.2 & 80.5 & 92.5 & 27.5 & 31.2 & 64.5 & 77.8 & 65.0 & 86.6 & 79.3 & 85.4 & 71.1 & 78.6 & 76.8 & 83.4 & 92.5 & 96.5 & 69.6 & 81.0 & 88.3 & 95.6 & 58.5 & 63.4 \\

\bottomrule
\end{tabular}
\end{sidewaystable}
\begin{sidewaystable}[p]
\centering
\caption{Detailed analysis of each category in nuScenes as the resolution changes ($\tau^*=5$ cm)}
\label{tab:nuscenes_resolution_detail}
\tiny 
\setlength{\tabcolsep}{1.2pt} 
\renewcommand{\arraystretch}{1.2} 
% Total columns: 2 (Scale, Method) + 3 (mIoU, mAcc, allAcc) + 16 categories * 2 (IoU, Acc) = 37 columns
\begin{tabular}{ll ccc *{16}{cc}}
\toprule
\textbf{Grid Size} & \textbf{Method} & \textbf{mIoU} & \textbf{mAcc} & \textbf{allAcc} & 
\multicolumn{2}{c}{Barrier} & \multicolumn{2}{c}{Bicycle} & \multicolumn{2}{c}{Bus} & 
\multicolumn{2}{c}{Car} & \multicolumn{2}{c}{Const.} & \multicolumn{2}{c}{Motor.} & 
\multicolumn{2}{c}{Ped.} & \multicolumn{2}{c}{Cone} & \multicolumn{2}{c}{Trailer} & 
\multicolumn{2}{c}{Truck} & \multicolumn{2}{c}{Drive.} & \multicolumn{2}{c}{Flat} & 
\multicolumn{2}{c}{Side.} & \multicolumn{2}{c}{Terr.} & \multicolumn{2}{c}{Man.} & 
\multicolumn{2}{c}{Veg.} \\

\cmidrule(lr){6-7} \cmidrule(lr){8-9} \cmidrule(lr){10-11} \cmidrule(lr){12-13} \cmidrule(lr){14-15} 
\cmidrule(lr){16-17} \cmidrule(lr){18-19} \cmidrule(lr){20-21} \cmidrule(lr){22-23} \cmidrule(lr){24-25} 
\cmidrule(lr){26-27} \cmidrule(lr){28-29} \cmidrule(lr){30-31} \cmidrule(lr){32-33} \cmidrule(lr){34-35} 
\cmidrule(lr){36-37}

& & (\%) & (\%) & (\%) & IoU & Acc & IoU & Acc & IoU & Acc & IoU & Acc & IoU & Acc & IoU & Acc & IoU & Acc & IoU & Acc & IoU & Acc & IoU & Acc & IoU & Acc & IoU & Acc & IoU & Acc & IoU & Acc & IoU & Acc & IoU & Acc \\
\midrule

% --- BLOCK 1: x0.5 ---
\multirow{3}{*}{$0.5\tau^*$} 
& Sp.Mink. & 34.1 & 41.4 & 85.7 & 18.3 & 18.8 & 1.4 & 1.4 & 21.3 & 46.2 & 40.5 & 41.2 & 9.6 & 21.4 & 1.2 & 1.2 & 9.2 & 9.5 & 11.7 & 11.9 & 3.2 & 7.8 & 19.9 & 34.6 & 93.1 & 98.4 & 34.1 & 37.2 & 58.4 & 66.0 & 68.1 & 85.2 & 75.5 & 89.9 & 79.8 & 92.0 \\
& PTv3     & 45.1 & 52.9 & 88.7 & 36.7 & 38.9 & 8.1 & 8.5 & 39.8 & 69.0 & 52.7 & 53.7 & 18.2 & 20.7 & 29.4 & 29.7 & 20.5 & 20.9 & 31.3 & 31.8 & 12.2 & 30.6 & 26.0 & 46.2 & 94.2 & 98.9 & 48.6 & 51.8 & 64.5 & 71.6 & 72.4 & 89.4 & 81.5 & 91.1 & 85.0 & 94.2 \\
& Invaria  & 74.3 & 82.2 & 93.6 & 77.0 & 85.3 & 25.5 & 64.8 & 90.5 & 91.8 & 90.7 & 93.9 & 44.1 & 49.4 & 82.8 & 84.3 & 75.7 & 79.3 & 63.7 & 69.1 & 63.3 & 73.5 & 80.7 & 87.1 & 96.5 & 98.5 & 72.5 & 80.3 & 75.0 & 84.7 & 75.2 & 85.4 & 88.4 & 94.3 & 86.7 & 94.4 \\
\midrule

% --- BLOCK 2: x1 ---
\multirow{3}{*}{$\tau^*$}
& Sp.Mink. & 76.6 & 82.0 & 93.7 & 77.6 & 85.8 & 37.1 & 40.7 & 91.7 & 93.1 & 86.9 & 89.0 & 52.9 & 57.1 & 84.6 & 86.8 & 80.1 & 83.7 & 64.0 & 69.3 & 69.4 & 74.6 & 86.5 & 91.8 & 96.7 & 98.5 & 73.0 & 81.9 & 75.6 & 85.3 & 75.2 & 85.0 & 88.8 & 94.9 & 86.1 & 94.1 \\
& PTv3     & 80.3 & 86.2 & 94.6 & 80.1 & 89.1 & 50.7 & 57.4 & 94.8 & 95.4 & 92.3 & 94.4 & 62.5 & 66.5 & 87.7 & 89.0 & 82.9 & 87.6 & 71.0 & 78.0 & 71.4 & 79.1 & 84.1 & 93.9 & 97.1 & 98.6 & 77.1 & 86.3 & 77.1 & 86.2 & 76.4 & 86.6 & 90.9 & 95.8 & 89.5 & 94.6 \\
& Invaria  & 76.2 & 83.9 & 94.0 & 78.2 & 86.7 & 26.9 & 66.5 & 92.3 & 93.0 & 91.8 & 95.0 & 49.9 & 54.9 & 83.7 & 85.0 & 77.6 & 81.3 & 65.2 & 71.8 & 69.3 & 77.0 & 83.8 & 89.3 & 96.7 & 98.4 & 75.2 & 83.6 & 75.6 & 85.2 & 75.5 & 85.8 & 89.3 & 94.9 & 87.7 & 94.3 \\
\midrule

% --- BLOCK 3: x2 ---
\multirow{3}{*}{$2\tau^*$} 
& Sp.Mink. & 35.4 & 45.7 & 85.1 & 23.0 & 28.2 & 0.3 & 0.3 & 0.3 & 0.3 & 43.9 & 47.5 & 7.7 & 52.4 & 4.6 & 7.4 & 12.6 & 17.8 & 29.0 & 55.2 & 6.6 & 10.4 & 24.6 & 30.6 & 92.6 & 94.6 & 45.9 & 51.7 & 56.8 & 73.5 & 61.8 & 78.4 & 74.7 & 95.7 & 81.8 & 87.2 \\
& PTv3     & 51.8 & 59.7 & 63.0 & 51.1 & 63.0 & 11.7 & 11.9 & 10.9 & 10.9 & 72.9 & 92.4 & 31.9 & 48.0 & 36.5 & 37.4 & 27.2 & 28.5 & 49.6 & 67.7 & 21.2 & 22.0 & 48.9 & 53.5 & 95.6 & 97.7 & 66.2 & 74.8 & 67.1 & 82.7 & 67.6 & 77.4 & 85.1 & 95.4 & 86.2 & 91.8 \\
& Invaria  & 75.6 & 83.4 & 94.0 & 76.8 & 83.9 & 25.9 & 64.5 & 91.7 & 92.5 & 90.8 & 94.6 & 50.6 & 54.6 & 84.1 & 85.4 & 76.5 & 80.7 & 63.8 & 71.7 & 66.6 & 75.2 & 83.5 & 89.0 & 96.7 & 98.3 & 74.6 & 84.1 & 75.2 & 85.1 & 74.9 & 85.8 & 89.4 & 95.3 & 88.2 & 94.1 \\
\midrule

% --- BLOCK 4: x3 ---
\multirow{3}{*}{$3\tau^*$} 
& Sp.Mink. & 14.5 & 22.6 & 58.8 & 8.7 & 9.6 & 0.0 & 0.0 & 0.0 & 0.0 & 0.2 & 0.2 & 3.4 & 18.1 & 0.0 & 0.0 & 1.1 & 1.5 & 11.9 & 24.7 & 0.7 & 0.8 & 0.8 & 0.8 & 55.9 & 58.6 & 2.6 & 2.6 & 19.1 & 29.3 & 24.8 & 65.3 & 51.8 & 97.5 & 51.1 & 53.1 \\
& PTv3     & 26.4 & 33.0 & 78.2 & 25.4 & 32.2 & 0.0 & 0.0 & 0.0 & 0.0 & 13.8 & 16.3 & 4.2 & 4.7 & 0.0 & 0.0 & 2.0 & 2.1 & 18.5 & 35.8 & 2.5 & 2.8 & 16.6 & 17.8 & 84.7 & 94.9 & 31.7 & 38.5 & 38.6 & 59.2 & 44.7 & 50.1 & 66.0 & 94.9 & 73.8 & 78.4 \\
& Invaria  & 71.9 & 80.1 & 93.0 & 73.9 & 82.9 & 20.2 & 53.1 & 89.8 & 90.5 & 89.0 & 94.4 & 48.8 & 52.4 & 77.1 & 79.9 & 68.3 & 72.6 & 56.4 & 66.2 & 61.3 & 68.6 & 79.2 & 84.7 & 96.1 & 98.1 & 72.0 & 83.3 & 71.6 & 82.6 & 72.1 & 84.2 & 88.0 & 94.5 & 86.4 & 92.8 \\

\bottomrule
\end{tabular}
\end{sidewaystable}
\begin{sidewaystable}[p] % This rotates the table 90 degrees on the page
\centering
\caption{Detailed analysis of each category in ScanNet as the scale changes. ($\tau^*=2 $cm)}
\label{tab:scale_details}
\tiny % Extremely small font to fit all columns
\setlength{\tabcolsep}{1.1pt} % Tighten column spacing
\renewcommand{\arraystretch}{1.2} % Vertical breathing room
\begin{tabular}{l l ccc *{20}{cc}}
\toprule
\textbf{Scale} & \textbf{Method} & \textbf{mIoU} & \textbf{mAcc} & \textbf{allAcc} & 
\multicolumn{2}{c}{Wall} & \multicolumn{2}{c}{Floor} & \multicolumn{2}{c}{Cab.} & 
\multicolumn{2}{c}{Bed} & \multicolumn{2}{c}{Chair} & \multicolumn{2}{c}{Sofa} & 
\multicolumn{2}{c}{Tabl} & \multicolumn{2}{c}{Door} & \multicolumn{2}{c}{Wind} & 
\multicolumn{2}{c}{Bksh} & \multicolumn{2}{c}{Pict} & \multicolumn{2}{c}{Cntr} & 
\multicolumn{2}{c}{Desk} & \multicolumn{2}{c}{Curt} & \multicolumn{2}{c}{Frid} & 
\multicolumn{2}{c}{Shw.} & \multicolumn{2}{c}{Toil} & \multicolumn{2}{c}{Sink} & 
\multicolumn{2}{c}{Bath} & \multicolumn{2}{c}{Oth.} \\

\cmidrule(lr){6-7} \cmidrule(lr){8-9} \cmidrule(lr){10-11} \cmidrule(lr){12-13} \cmidrule(lr){14-15} \cmidrule(lr){16-17} \cmidrule(lr){18-19} \cmidrule(lr){20-21} \cmidrule(lr){22-23} \cmidrule(lr){24-25} \cmidrule(lr){26-27} \cmidrule(lr){28-29} \cmidrule(lr){30-31} \cmidrule(lr){32-33} \cmidrule(lr){34-35} \cmidrule(lr){36-37} \cmidrule(lr){38-39} \cmidrule(lr){40-41} \cmidrule(lr){42-43} \cmidrule(lr){44-45}

& & (\%) & (\%) & (\%) & IoU & Acc & IoU & Acc & IoU & Acc & IoU & Acc & IoU & Acc & IoU & Acc & IoU & Acc & IoU & Acc & IoU & Acc & IoU & Acc & IoU & Acc & IoU & Acc & IoU & Acc & IoU & Acc & IoU & Acc & IoU & Acc & IoU & Acc & IoU & Acc & IoU & Acc & IoU & Acc \\
\midrule

% --- BLOCK 1: x2 ---
\multirow{7}{*}{$\times 2$} 

& Sonata & 57.9 & 69.0 & 84.1 & 81.2 & 94.9 & 94.4 & 97.9 & 49.3 & 67.9 & 61.6 & 80.4 & 75.5 & 78.3 & 56.2 & 80.5 & 56.0 & 74.4 & 50.0 & 59.2 & 59.4 & 72.3 & 64.2 & 75.4 & 37.2 & 47.6 & 46.1 & 66.5 & 30.0 & 35.6 & 75.5 & 85.1 & 31.9 & 39.0 & 66.8 & 71.9 & 59.8 & 61.2 & 49.6 & 58.0 & 74.4 & 82.8 & 39.3 & 51.2 \\
& Concrt-L & 65.9 & 75.5 & 87.8 & 84.1 & 95.4 & 94.7 & 97.8 & 61.7 & 80.0 & 72.1 & 87.0 & 79.3 & 82.8 & 62.0 & 82.8 & 63.9 & 77.1 & 56.5 & 62.4 & 71.3 & 84.0 & 76.4 & 83.8 & 32.5 & 37.4 & 57.7 & 74.6 & 49.4 & 57.9 & 73.7 & 88.9 & 60.8 & 67.2 & 41.0 & 42.9 & 81.1 & 82.8 & 61.1 & 68.1 & 82.8 & 86.5 & 56.8 & 69.9 \\
& Utonia   & 59.0 & 69.2 & 83.6 & 77.5 & 95.5 & 94.0 & 97.0 & 52.3 & 66.8 & 67.4 & 82.8 & 73.3 & 76.3 & 57.9 & 83.9 & 53.1 & 68.8 & 48.9 & 54.6 & 57.3 & 69.9 & 62.6 & 68.2 & 30.8 & 36.6 & 51.7 & 74.1 & 35.5 & 39.6 & 66.7 & 83.9 & 52.4 & 55.3 & 59.9 & 64.0 & 71.9 & 72.5 & 54.6 & 64.3 & 70.3 & 73.3 & 42.0 & 56.3 \\
& Sp.Mink. & 35.7 & 46.3 & 73.2 & 72.8 & 95.4 & 92.4 & 98.3 & 30.6 & 50.8 & 40.9 & 81.0 & 36.0 & 37.1 & 23.0 & 42.6 & 40.2 & 59.1 & 25.8 & 30.4 & 39.0 & 49.1 & 40.6 & 45.5 & 19.3 & 24.1 & 19.1 & 24.2 & 16.2 & 20.4 & 54.7 & 70.5 & 9.3 & 11.0 & 32.3 & 33.9 & 28.5 & 30.2 & 23.7 & 24.5 & 47.8 & 63.6 & 21.1 & 34.4 \\
& PTv3 & 49.5 & 58.4 & 80.1 & 76.2 & 96.7 & 89.7 & 98.9 & 44.8 & 64.9 & 58.9 & 80.1 & 69.2 & 72.1 & 38.9 & 54.8 & 46.9 & 69.6 & 39.2 & 44.8 & 47.4 & 55.9 & 58.0 & 61.1 & 23.1 & 25.2 & 24.5 & 26.8 & 24.2 & 26.0 & 60.5 & 75.3 & 34.0 & 36.6 & 46.5 & 49.5 & 61.4 & 62.7 & 47.6 & 52.4 & 68.9 & 80.9 & 29.3 & 34.6 \\
& Invaria & 75.6 & 83.8 & 91.1 & 86.5 & 95.0 & 95.6 & 98.4 & 70.8 & 81.8 & 83.5 & 88.9 & 91.5 & 95.5 & 84.2 & 94.1 & 74.2 & 82.0 & 68.1 & 77.5 & 68.3 & 81.4 & 82.6 & 92.2 & 37.8 & 43.0 & 65.3 & 75.6 & 65.7 & 89.6 & 79.4 & 86.1 & 74.5 & 81.7 & 76.4 & 83.0 & 95.3 & 97.2 & 69.7 & 78.7 & 88.9 & 94.1 & 60.6 & 66.5 \\
\midrule

% --- BLOCK 2: x1 ---
\multirow{7}{*}{$\times 1$} 
& Sonata   & 72.6 & 83.4 & 89.8 & 85.7 & 93.0 & 95.1 & 97.8 & 63.2 & 75.8 & 78.5 & 84.8 & 88.0 & 93.2 & 79.3 & 91.0 & 71.5 & 81.7 & 67.7 & 81.1 & 68.2 & 84.4 & 77.2 & 92.4 & 37.6 & 49.3 & 63.7 & 80.3 & 62.7 & 86.0 & 81.5 & 87.8 & 57.0 & 74.8 & 73.2 & 80.2 & 91.2 & 96.9 & 67.7 & 80.8 & 85.8 & 94.1 & 55.0 & 62.2 \\
& Concrt-L & 78.6 & 87.4 & 92.3 & 87.4 & 93.0 & 95.5 & 97.7 & 72.3 & 82.7 & 82.3 & 89.9 & 90.7 & 95.5 & 81.8 & 90.7 & 78.5 & 86.8 & 77.5 & 91.7 & 79.1 & 89.4 & 86.7 & 94.2 & 37.6 & 53.7 & 68.8 & 85.6 & 68.6 & 80.5 & 86.0 & 91.1 & 79.8 & 89.9 & 74.9 & 80.3 & 94.0 & 96.6 & 72.0 & 82.8 & 86.9 & 94.3 & 71.9 & 81.8 \\
& Utonia   & 76.6 & 85.1 & 94.6 & 85.8 & 94.6 & 95.4 & 97.9 & 72.0 & 82.0 & 81.0 & 87.1 & 88.8 & 91.8 & 73.8 & 91.0 & 77.0 & 83.8 & 75.1 & 85.4 & 73.1 & 82.8 & 82.7 & 92.4 & 39.0 & 50.5 & 65.2 & 86.3 & 67.9 & 79.0 & 83.3 & 88.0 & 80.0 & 86.6 & 75.1 & 80.7 & 94.0 & 96.0 & 70.2 & 79.5 & 85.3 & 90.4 & 66.6 & 75.8 \\
& Sp.Mink. & 75.6 & 83.5 & 91.5 & 87.2 & 95.7 & 95.9 & 98.2 & 69.9 & 80.6 & 84.6 & 89.9 & 92.4 & 95.7 & 84.7 & 93.3 & 76.2 & 84.9 & 70.0 & 80.2 & 68.4 & 81.5 & 81.7 & 92.2 & 37.8 & 44.2 & 66.0 & 80.9 & 68.1 & 86.1 & 77.9 & 86.4 & 65.9 & 72.8 & 71.8 & 76.4 & 91.0 & 96.6 & 67.5 & 72.1 & 88.8 & 91.4 & 65.6 & 70.1 \\
& PTv3 & 77.5 & 85.0 & 92.0 & 87.2 & 95.7 & 95.7 & 98.3 & 71.2 & 83.1 & 83.4 & 87.8 & 92.9 & 96.4 & 86.3 & 93.6 & 79.0 & 86.6 & 74.9 & 84.4 & 74.0 & 84.1 & 83.0 & 91.6 & 39.8 & 48.5 & 71.7 & 81.7 & 73.9 & 89.0 & 79.1 & 87.5 & 71.3 & 75.7 & 72.2 & 77.3 & 93.4 & 97.9 & 71.0 & 81.5 & 88.2 & 91.4 & 63.0 & 66.7 \\
& Invaria & 76.2 & 84.3 & 91.2 & 86.6 & 94.9 & 95.7 & 98.4 & 71.2 & 82.0 & 83.6 & 89.0 & 91.9 & 95.7 & 84.6 & 93.9 & 74.2 & 82.4 & 69.1 & 78.7 & 69.9 & 82.6 & 83.1 & 92.6 & 38.4 & 44.0 & 65.5 & 76.6 & 65.4 & 89.3 & 79.8 & 86.7 & 75.5 & 83.2 & 74.2 & 80.0 & 95.7 & 97.8 & 69.8 & 78.3 & 89.6 & 94.2 & 60.1 & 66.1 \\
\midrule

% --- BLOCK 3: x1/2 ---
\multirow{7}{*}{$\times 1/2$} 
& Sonata   & 60.7 & 72.4 & 85.1 & 80.2 & 90.0 & 93.2 & 98.2 & 52.2 & 64.8 & 71.0 & 82.5 & 78.4 & 87.9 & 68.4 & 75.0 & 62.7 & 71.6 & 54.0 & 68.4 & 59.9 & 77.0 & 69.8 & 89.5 & 21.8 & 26.4 & 48.5 & 53.7 & 50.2 & 82.0 & 75.1 & 84.9 & 41.6 & 56.8 & 67.2 & 76.2 & 63.5 & 70.1 & 47.3 & 64.3 & 68.4 & 78.1 & 39.5 & 51.1 \\
& Concrt-L & 71.5 & 81.9 & 89.5 & 83.3 & 90.0 & 93.9 & 97.9 & 65.9 & 77.7 & 80.6 & 89.2 & 85.3 & 91.7 & 82.4 & 86.8 & 72.5 & 79.2 & 66.5 & 83.5 & 72.6 & 87.3 & 81.4 & 95.6 & 31.8 & 49.4 & 60.2 & 71.4 & 62.4 & 80.2 & 82.5 & 90.0 & 66.0 & 80.0 & 72.6 & 80.3 & 68.1 & 69.8 & 59.9 & 74.3 & 79.4 & 87.6 & 62.1 & 75.9 \\
& Utonia   & 71.8 & 82.5 & 89.4 & 83.2 & 91.5 & 94.0 & 98.0 & 67.8 & 80.1 & 81.6 & 87.8 & 85.2 & 91.1 & 79.2 & 89.2 & 70.8 & 75.8 & 67.8 & 81.2 & 70.0 & 84.6 & 81.2 & 93.6 & 31.1 & 41.9 & 61.0 & 74.9 & 63.9 & 81.5 & 77.7 & 86.3 & 63.3 & 80.0 & 72.9 & 82.5 & 85.5 & 93.1 & 59.6 & 76.3 & 79.5 & 90.4 & 60.4 & 71.2 \\
& Sp.Mink. & 46.9 & 55.4 & 82.1 & 78.0 & 95.0 & 93.8 & 98.3 & 44.2 & 60.9 & 56.5 & 59.1 & 74.8 & 84.1 & 53.8 & 61.0 & 59.5 & 76.3 & 38.2 & 48.0 & 55.3 & 71.5 & 69.0 & 80.6 & 16.1 & 17.4 & 45.0 & 54.0 & 41.0 & 53.5 & 64.6 & 79.8 & 24.3 & 25.2 & 33.8 & 35.2 & 10.2 & 10.6 & 28.3 & 31.1 & 23.1 & 25.0 & 28.7 & 40.8 \\
& PTv3 & 59.4 & 68.3 & 85.7 & 80.4 & 95.3 & 92.8 & 98.7 & 58.2 & 71.1 & 67.0 & 70.8 & 80.1 & 88.0 & 56.8 & 66.7 & 67.5 & 79.5 & 52.1 & 60.0 & 61.7 & 76.8 & 77.6 & 86.3 & 27.3 & 31.8 & 57.2 & 64.5 & 44.2 & 56.4 & 72.0 & 82.3 & 39.2 & 40.8 & 49.4 & 52.9 & 51.0 & 51.3 & 53.4 & 65.7 & 58.5 & 77.2 & 41.0 & 49.7 \\
& Invaria & 75.6 & 84.0 & 90.8 & 85.9 & 94.3 & 95.3 & 98.5 & 68.9 & 81.5 & 83.3 & 86.7 & 91.5 & 95.2 & 83.9 & 94.9 & 74.2 & 82.2 & 65.4 & 75.9 & 68.1 & 82.9 & 83.3 & 93.0 & 34.5 & 40.1 & 66.1 & 78.3 & 65.4 & 88.5 & 79.4 & 86.1 & 75.4 & 81.7 & 76.4 & 83.0 & 95.3 & 97.2 & 69.7 & 78.7 & 88.9 & 94.1 & 61.8 & 66.8 \\
\midrule
% --- BLOCK 4: x1/3 ---
\multirow{7}{*}{$\times 1/3$} 
& Sonata   & 32.1 & 41.6 & 70.9 & 70.8 & 91.7 & 87.0 & 95.7 & 24.9 & 34.3 & 39.0 & 51.0 & 41.4 & 48.8 & 16.5 & 17.5 & 39.3 & 52.4 & 21.8 & 27.2 & 45.2 & 59.8 & 48.1 & 72.9 & 4.8 & 4.9 & 4.7 & 9.3 & 19.4 & 29.6 & 60.1 & 68.7 & 15.1 & 19.1 & 52.0 & 64.2 & 8.4 & 10.5 & 11.6 & 18.1 & 20.4 & 29.0 & 12.1 & 27.3 \\
& Concrt-L & 43.3 & 55.3 & 77.1 & 74.6 & 86.4 & 88.9 & 97.2 & 36.7 & 59.0 & 62.7 & 74.4 & 55.6 & 62.2 & 57.4 & 61.7 & 49.6 & 62.9 & 39.6 & 54.1 & 56.9 & 80.1 & 46.3 & 69.4 & 21.1 & 40.7 & 17.3 & 25.4 & 30.4 & 52.5 & 69.3 & 80.5 & 25.8 & 28.3 & 59.1 & 64.5 & 4.2 & 4.3 & 18.0 & 29.9 & 30.9 & 36.6 & 21.4 & 35.3 \\
& Utonia   & 53.7 & 65.7 & 81.1 & 76.7 & 88.8 & 90.4 & 97.2 & 47.2 & 64.5 & 73.8 & 79.2 & 63.4 & 71.0 & 69.3 & 81.1 & 53.5 & 63.3 & 47.4 & 61.9 & 57.8 & 78.3 & 53.9 & 65.1 & 16.5 & 22.6 & 30.1 & 33.8 & 41.6 & 54.9 & 69.2 & 84.9 & 45.4 & 59.5 & 67.4 & 76.1 & 34.1 & 36.4 & 38.4 & 60.6 & 66.1 & 79.2 & 32.2 & 54.8 \\
& Sp.Mink. & 19.1 & 25.4 & 65.6 & 64.7 & 95.0 & 84.6 & 95.3 & 20.0 & 31.1 & 7.0 & 7.1 & 33.5 & 38.3 & 8.7 & 8.8 & 35.8 & 54.1 & 8.8 & 11.0 & 32.4 & 51.4 & 22.5 & 25.8 & 0.8 & 0.8 & 0.0 & 0.0 & 5.6 & 7.2 & 45.6 & 56.0 & 0.0 & 0.0 & 1.0 & 1.0 & 0.0 & 0.0 & 0.7 & 0.7 & 0.0 & 0.0 & 9.6 & 23.3 \\
& PTv3 & 20.1 & 25.6 & 65.4 & 60.0 & 96.1 & 74.0 & 98.9 & 12.8 & 14.8 & 10.1 & 10.2 & 28.4 & 32.5 & 2.9 & 2.9 & 35.0 & 42.9 & 18.6 & 21.2 & 40.1 & 60.8 & 24.4 & 25.5 & 5.5 & 7.3 & 0.0 & 0.0 & 2.3 & 2.4 & 49.9 & 53.8 & 0.0 & 0.0 & 12.8 & 13.1 & 0.0 & 0.0 & 4.7 & 5.0 & 11.2 & 12.1 & 8.6 & 11.5 \\
& Invaria & 73.8 & 82.7 & 90.0 & 84.5 & 94.0 & 95.1 & 98.5 & 67.8 & 80.8 & 82.1 & 86.8 & 90.2 & 94.2 & 82.7 & 95.4 & 73.6 & 81.3 & 59.8 & 70.4 & 65.9 & 80.3 & 80.5 & 92.5 & 27.6 & 31.3 & 64.8 & 77.9 & 64.9 & 86.6 & 79.4 & 85.5 & 71.2 & 78.8 & 76.8 & 83.5 & 92.7 & 96.5 & 69.6 & 81.1 & 88.7 & 95.5 & 58.4 & 63.3 \\

\bottomrule
\end{tabular}
\end{sidewaystable}

\newpage
\section{More Visualization}
\label{sec:more_vis}
\begin{figure} % 't' helps placement at the top of the page
\captionsetup{skip=3pt}
   \centering
   \includegraphics[width=0.9\linewidth]{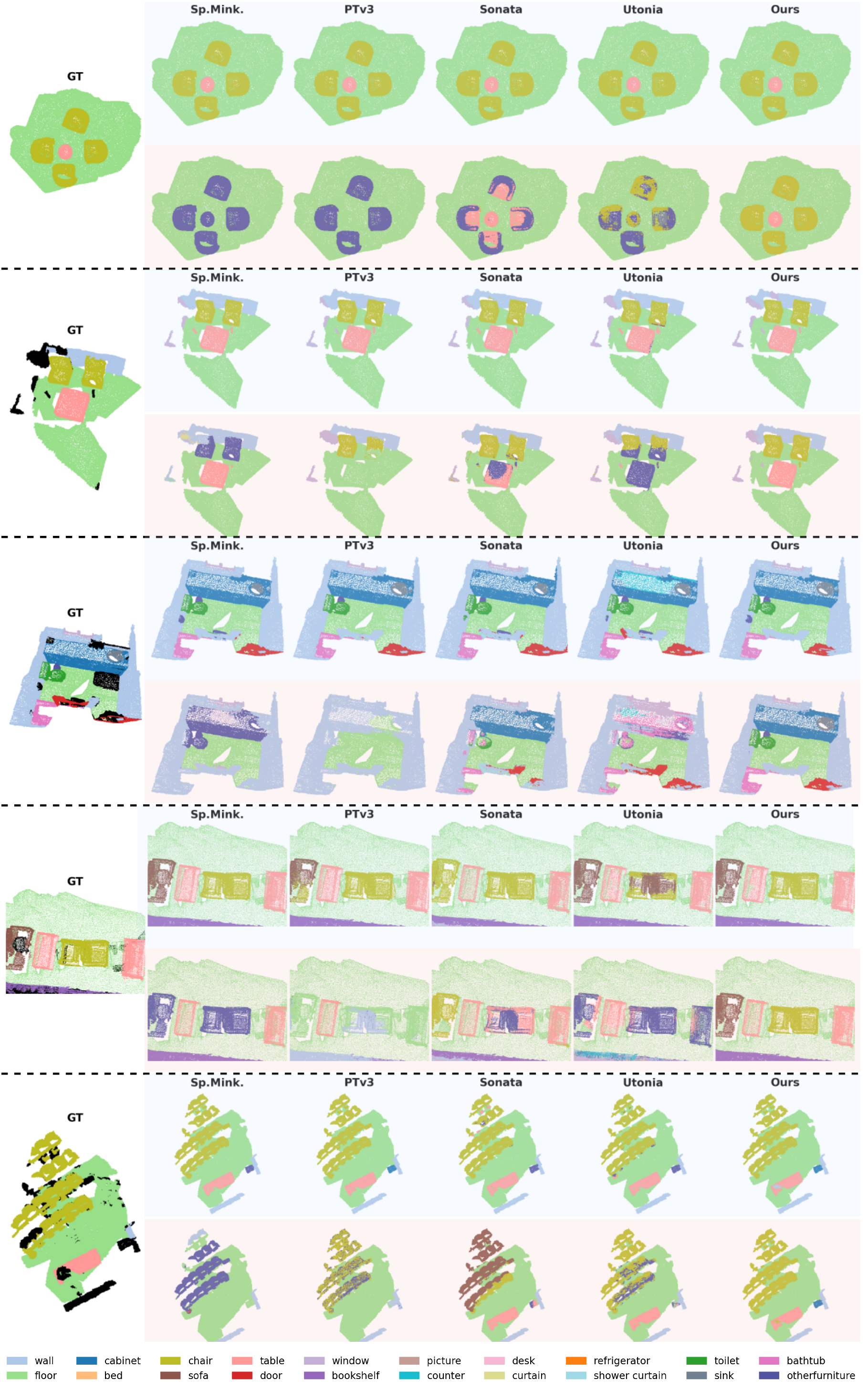}
   \caption{ Comparison on ScanNet semantic segmentation. The first row displays model performance
when evaluated at the same resolution. The second row illustrates the impact of shifting the resolution
during inference. For clear visual comparison, results from both are projected on the original point
cloud. The black spot in GT means no label. }
   \label{fig:qualitative_more}
\end{figure}

In \Cref{fig:qualitative_more}, we provide additional qualitative comparisons between our method and existing approaches, including Sp.Mink.~\cite{choy20194d}, PTv3~\cite{wu2024point}, and Utonia~\cite{zhang2026utonia}.
%%%%%%%%%%%%%%%%%%%%%%%%%%%%%%%%%%%%%%%%%%%%%%%%%%%%%%%%%%%%

% NeurIPS paper checklist: needed for the camera-ready, left out of the arXiv version.
%\clearpage
%\input{checklist}

\end{document}